\documentclass[a4paper,10pt,conference,doublespace]{IEEEtran}

\makeatletter
\let\NAT@parse\undefined
\makeatother

\usepackage[dvipsnames]{xcolor}
\usepackage[pdftex]{graphicx}
\newcommand*\linkcolours{ForestGreen}
\usepackage{times}
\usepackage{graphicx}
\usepackage{amssymb}
\usepackage{gensymb}
\usepackage{amsmath}
\usepackage{multirow}
\usepackage{breakurl}

\usepackage{url,hyperref}
\hypersetup{
colorlinks,
linkcolor=\linkcolours,
citecolor=\linkcolours,
filecolor=\linkcolours,
urlcolor=\linkcolours}

\usepackage{algorithm}
\usepackage{algorithmic}

\usepackage[labelfont={bf},font=small]{caption}
\usepackage[none]{hyphenat}

\begin{document}
\title{JobRecoGPT: Explainable job recommendations using LLMs}

\author{
\authorblockN{Preetam Ghosh, Vaishali Sadaphal}
\authorblockA{Intelligent Service Networks, Tata Research, Tata Consultancy Services.}
}

\maketitle

%-------------------------------------------------------------------------
\begin{abstract}
In today's rapidly evolving job market, finding the right opportunity can be a daunting challenge. With advancements in the field of AI, computers can now recommend suitable jobs to candidates. However, the task of recommending jobs is not same as recommending movies to viewers. Apart from must-have criteria, like skills and experience, there are many subtle aspects to a job which can decide if it is a good fit or not for a given candidate. Traditional approaches can capture the quantifiable aspects of jobs and candidates, but a substantial portion of the data that is present in unstructured form in the job descriptions and resumes is lost in the process of conversion to structured format. As of late, Large Language Models (LLMs) have taken over the AI field by storm with extraordinary performance in fields where text-based data is available. Inspired by the superior performance of LLMs, we leverage their capability to understand natural language for capturing the information that was previously getting lost during the conversion of unstructured data to structured form. To this end, we compare performance of four different approaches for job recommendations namely, (i) Content based deterministic, (ii) LLM guided, (iii) LLM unguided, and (iv) Hybrid. In this study, we present advantages and limitations of each method and evaluate their performance in terms of time requirements.
\end{abstract}
%-------------------------------------------------------------------------
%%%%%%%%%%%%%%%%%%%%%%%%%%%%%%%%%%%%%%%%%%%%%%%%%%%%%%%%%%
\section{Introduction}
%%%%%%%%%%%%%%%%%%%%%%%%%%%%%%%%%%%%%%%%%%%%%%%%%%%%%%%%%%
Identifying job opportunities for talent is important to enable organisations to attract, develop, and retain the talent.
It is a time-consuming process when done manually and results in limited reach, inconsistent criteria with human bias.
With the emergence of freelance platforms %\cite{toptal, dribbble}
%, upworkinc, fiverr, taskrabbit, designcrowd, gigster},
and integration of technology, significance of data-driven job recommendations has grown.

Traditional state-of-the-art techniques 
%\cite{de2021job,kenthapadi2017personalized,yang2017combining}%\cite{job_reco_algo1,de2021job,kenthapadi2017personalized,yang2017combining}
recommend job opportunities to the talent based on similarity between job requirements and talent attributes.
However, the nature of data in this domain is inherently in unstructured natural language format viz. resume and job descriptions.
To use traditional approaches, one is required to extract information and bring it to a structured format.
In this study, we investigate the application of large language models (LLMs) \cite{zhao2023survey} due to their ability to process and comprehend language, as well as their extensive knowledge gained from training on Internet text.

Providing recommendations falls in the category of Reasoning intelligence, specifically Prescriptive intelligence.
Generative AI and language models are observed to be performing well in Recognition and Operative intelligence. In this study, we investigate role of language models in Reasoning intelligence.

The traditional methods are required to define a structure that consist of requirements of jobs and corresponding attributes of talent such as role, skills, educational background, and experience, among others.
The jobs with closest match with the attributes of talent are recommended.
These methods face several challenges.
\begin{itemize}
\item The task of transforming data from CVs and JDs into a structured format is error-prone and can result in loss of information \cite{data_extraction}.
%\cite{scikit_learn, spacy, data_extraction}.
\item The qualitative aspects of talent such as achievements, strengths, and aspirations are ambiguous and presented in natural language, hence not extracted.
\item The data driven tools rely on quantitative metrics, such as skills and experience, while overlooking qualitative aspects like soft skills or potential for growth.
\item The JDs may be incomplete, the qualitative aspects of the job and organization may be biased or even may not be mentioned.
\end{itemize}

In this work, we leverage the capability of LLM's to
understand natural language for capturing the information that
was previously getting lost during the conversion of unstructured
data to structured form. We present,
\begin{itemize}
\item One content based deterministic approach: based on traditional techniques. This is used as a baseline to compare performance of all approaches
\item Two LLM based approaches: guided and unguided
\item A hybrid approach: a combination of traditional and LLM based approach
\item Evaluation: Comparison of quality of recommendations produced by each method and the efficiency.
\end{itemize}

Though the approaches proposed in this work are generic, we
consider the domain of Information Technology to evaluate the effectiveness of these techniques. We conduct experiments using two datasets:
\begin{itemize}
\item \emph{Synthetic data}: In this experiment, we generate synthetic data that simulates the characteristics of the IT domain, allowing us to assess the performance of all methods in a controlled environment.
\item \emph{Real world data}: In this experiment, we use real JDs from the IT field. This enables us to evaluate the practical applicability and performance of the  methods using authentic data.
\end{itemize}
Through these experiments, we aim to gain insights into the effectiveness and limitations of the all the techniques for providing ranked job recommendations for a talent in the IT domain.

%%%%%%%%%%%%%%%%%%%%%%%%%%%%%%%%%%%%%%%%%%%%%%%%%%%%%%%%%%
\section{Content based Deterministic approach} 
\label{sec:deterministic_algorithm}
%%%%%%%%%%%%%%%%%%%%%%%%%%%%%%%%%%%%%%%%%%%%%%%%%%%%%%%%%%
Deterministic approach is on the lines of content based techniques of recommending jobs to talents.
This simple approach is used as a baseline to compare performance of all approaches.
In this approach, a job is recommended by matching its requirements with the attributes of talent.

\begin{figure}
\centering
\includegraphics[width = 3in, height = 1.2in]{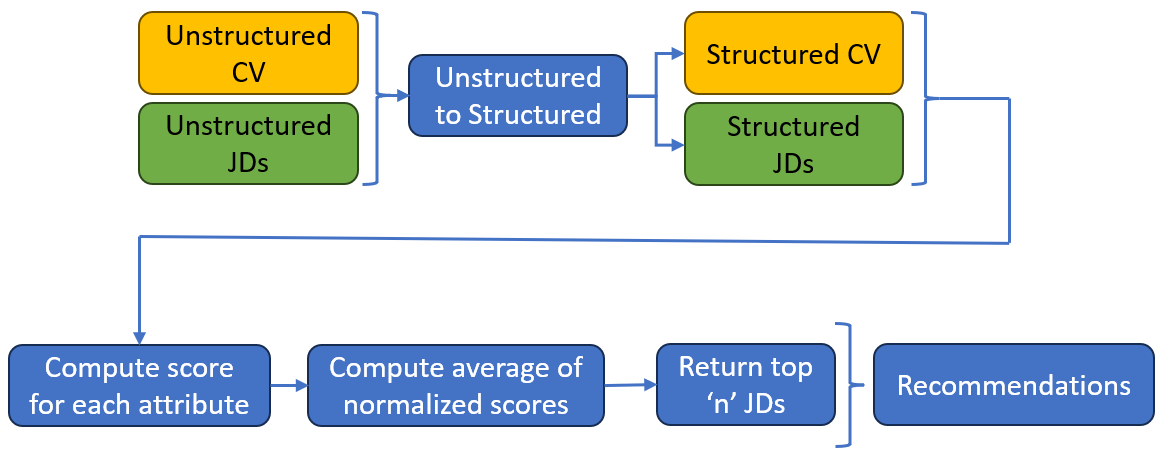}
\caption{Deterministic algorithm.}
\label{fig:deterministic_algorithm}
\end{figure}

Refer Figure \ref{fig:deterministic_algorithm}. The algorithm accepts inputs as unstructured resume (CV) and job descriptions (JDs) and configurations that include objective direction, indicating if higher, lower or closer values of the job attributes with respect to the talent attribute is better and the number of recommendations required.
The output is the recommended JDs with a score.

\subsubsection{Unstructured to structured conversion}

\begin{figure}
\centering
\includegraphics[width = 3.5in, height = 4.2in]{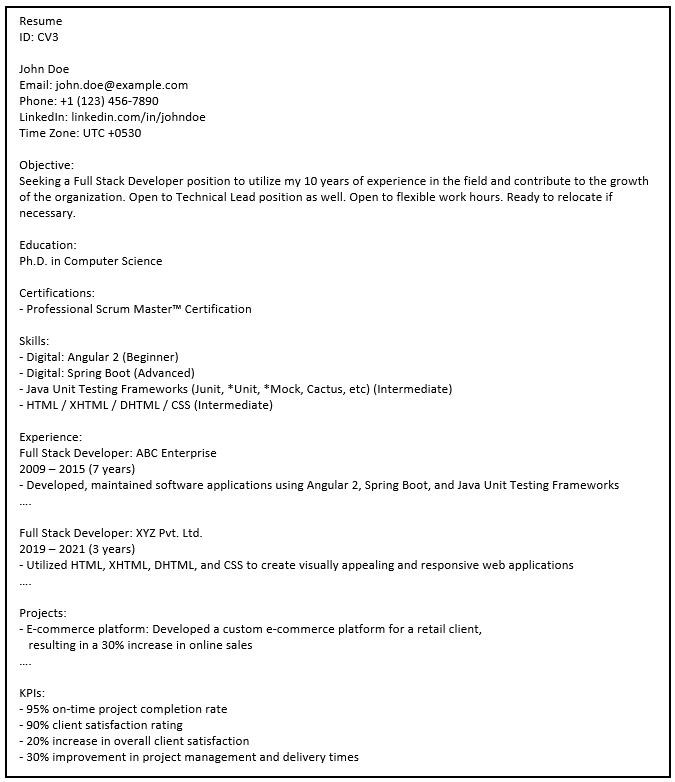}
\caption{Unstructured Resume CV3.}
\label{fig:unstructured_resume_for_cv3}
\end{figure}

\begin{figure}
\centering
\includegraphics[width = 2in, height = 0.9in]{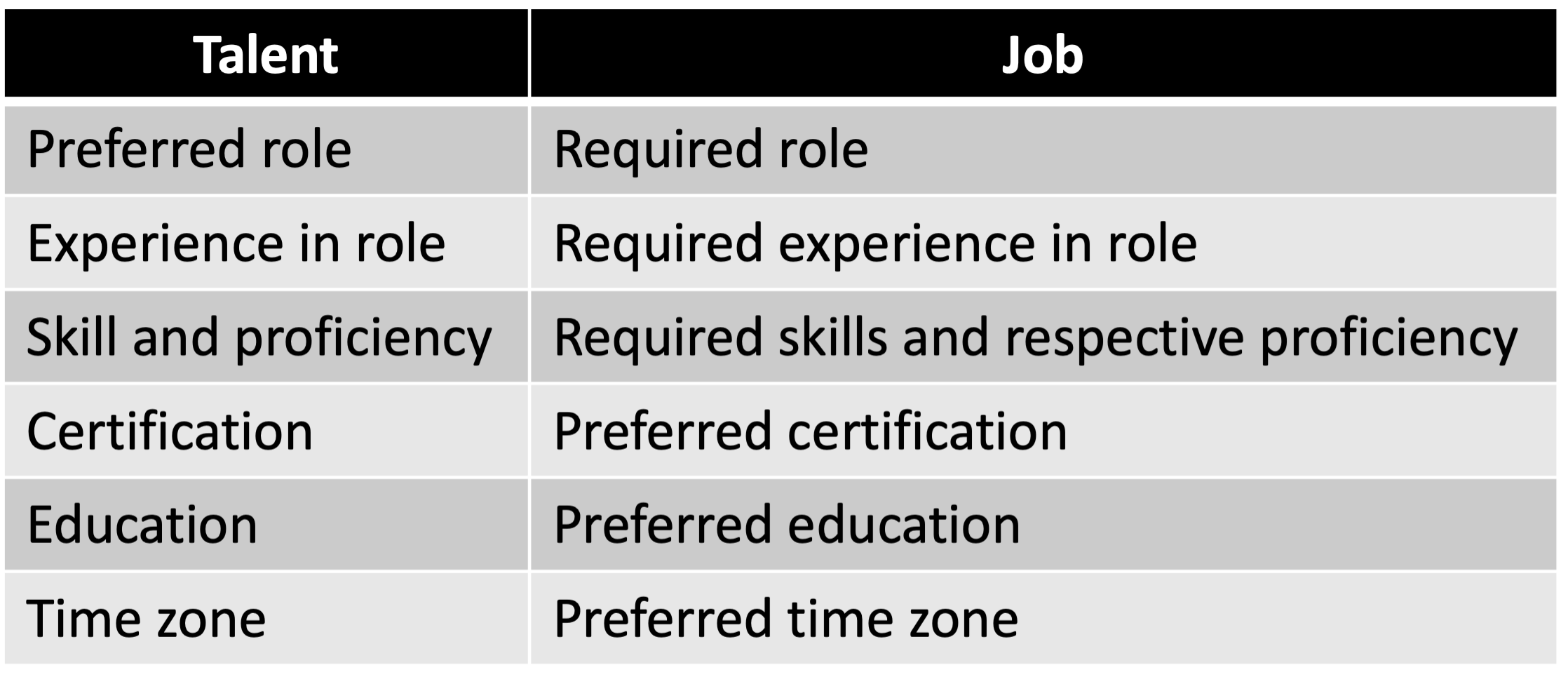}
\caption{Attributes of talent and job.}
\label{fig:talent_job_attributes}
\end{figure}

Figure \ref{fig:unstructured_resume_for_cv3} and \ref{fig:unstructured_five_jds_for_cv3} show a CV and a JD, respectively.
Figure \ref{fig:talent_job_attributes} shows the attribute model for a talent and a job.
The talent attributes capture talent's professional information and preferences. Whereas, the job attributes capture corresponding requirements.

\begin{figure}
\centering
\includegraphics[width = 3.0in, height = 2.7in]{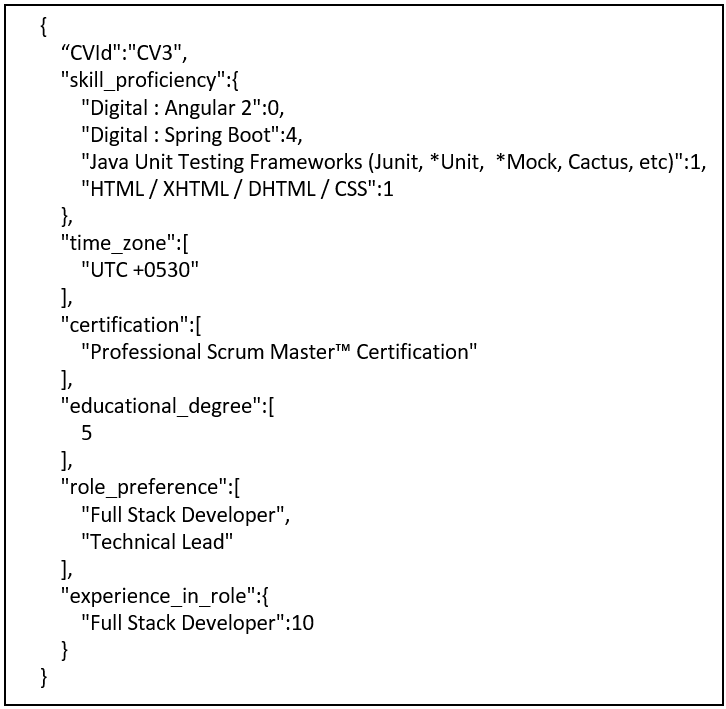}
\caption{Structured resume CV3.}
\label{fig:structured_resume_for_cv3}
\end{figure}

The talent and job model is populated by extracting information from the CV and JD.
The structured model of talent and job corresponding to the CV and JD in
Figure \ref{fig:unstructured_resume_for_cv3} and \ref{fig:unstructured_five_jds_for_cv3}
is shown in
Figure \ref{fig:structured_resume_for_cv3}, and \ref{fig:structured_five_jds_for_cv3} respectively.
%The talent is a Ph.D. holder with 10 years experience as a ``Full Stack Developer" with suitable skills and certifications as mentioned in the resume and a preferred time zone of work.
%Similarly, the requirement of the job is for role the role of ``Technical Lead", the corresponding skill, educational requirements are captured in the attributes.
%Many tools are available to extract information from unstructured text \cite{data_extraction}.
We use LLMs to convert unstructured data to structured form. The prompts used for this conversion are depicted in Figure \ref{fig:synthetic_data_generation_prompt}.

\subsubsection{Deterministic algorithm}

The deterministic algorithm, as the name suggests, generates predictable, hence reproducible results every time, based on the criteria provided to it. It provides a baseline score to compare recommendations from other approaches. Though there are complex recommendation approaches, we keep it simple while capturing the required attributes that determine if a job is suitable for a given talent.

At the core of this algorithm, are basic comparisons between various attributes extracted from the CV and JD.
%We use two types of matches between the attributes: ``closer” and ``exact”. 

\begin{itemize}
    \item ``Closer" match:
    %Here, we assess the extent to which the attribute value in the CV deviates from the specified values in the JD.
    A rating of 1 is awarded with absolute deviation equal to zero and the rating decreases as absolute deviation increases.
    %= 0 and  scores decreasing proportionally in both directions for values higher or lower than specified. 
    This approach penalizes both under-qualified candidates. It also penalizes over-qualified candidates as they may be better suited for other roles.
    
    \item ``Exact" match: A score of 1 is awarded if a value or a set are an exact match. This is important for some attributes where we a specific value is required, say mandatory certification.
\end{itemize}
For our experiment, we configure all the attributes set to ``closer” match, the only exception is the ``certifications” for which we set it to “exact”.
Consider the following example.
%of a resume CV3 and job description JD9 in Figure \ref{fig:structured_resume_for_cv3} and ref{}. JD9 has a matching score of 0.77 as explained below.

\begin{enumerate}
    \item Skill proficiency: Consider that three skills out of four are common, with one having same proficiency and the other two with a deviation of 2 from required. This leads to a score: $(1 + 1/2 + 1/2)/4 = 0.5$. The first value is 1 as the proficiency exactly matches with the requirement and the next two values have $1/2$ as the values deviate by 2. %Finally, this is divided by the required number of skills, 4.
    \item Time zone: It is an an absolute difference in hours and is calculated ranging between 0 to 26 which is then normalized between 1 and 0.
    \item Certification: It is a set match of the required certifications. The score is 1 for a complete set match, else 0.
    \item Education: It is denoted as a number from 1 to 5 with higher value for higher education. Its score is the reciprocal of the absolute difference with respect to required value.
    \item Experience: The score is calculated as a ratio of talent value and required value in the required role. For example, for the matching role the actual value is 6 and required is 10, resulting in score of $(6/10) = 0.6$.
    \item Role: It is the reciprocal of the preference of the required role by the talent. For example, the candidate has 2nd preference for the role, resulting in a score of 0.5.
    \item Finally, an average value of all scores is  $(0.5+1+1+1+0.6+0.5)/6 = 4.6/6 = 0.77$.
\end{enumerate}

%The overall flow of the algorithm is presented in Figure \ref{fig:deterministic_algorithm}.

%%%%%%%%%%%%%%%%%%%%%%%%%%%%%%%%%%%%%%%%%%%%%%%%%%%%%%%%%%
\section{Language model based approach}
%%%%%%%%%%%%%%%%%%%%%%%%%%%%%%%%%%%%%%%%%%%%%%%%%%%%%%%%%%

Though large language models (LLMs) are predictors of next words and lack true comprehension or knowledge, they have a convincing ability to generate coherent responses and recall information. This makes it seem like they possess knowledge of many domains. We propose to leverage this ability of LLMs to understand and correlate different required skills and roles based on their similarity. For example, someone experienced in design engineering and software development in the domain of IT may inherently be suited for a Full stack developer role, even if not explicitly mentioned. This provides a significant advantage over methods that are constrained by specific value and text based similarities.

We provide unstructured CV and JDs to a language model to generate job recommendations.
Refer Figure \ref{fig:llm_guided_algorithm} and \ref{fig:llm_unguided_algorithm} for the two LLM based approaches.
We use GPT4 \cite{openai_gpt4} model of OpenAI for this purpose.

\subsection{LLM Guided Algorithm}
Here, we provide a pre-defined structure to generate recommendations. This allows control over the model's response, such as the criteria used for matching, the number of recommendations, and returning recommendations in a structured format.
Further, we exploit LLMs ability to explain their actions in natural language to provide reasoning of why a certain recommendation is good or bad.

\begin{figure}
\centering
\includegraphics[width = 3in, height = 3.4in]{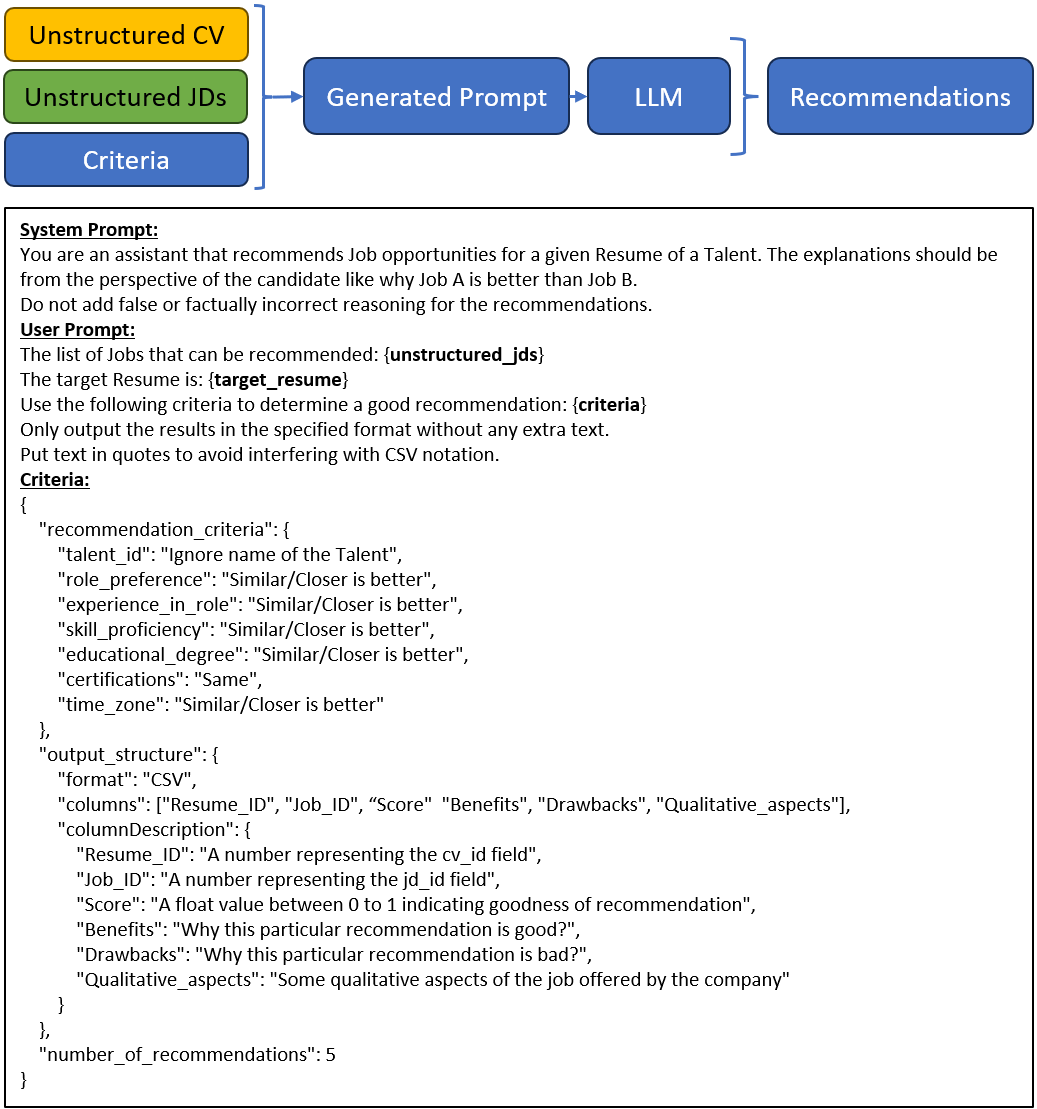}
\caption{Language model based guided algorithm.}
\label{fig:llm_guided_algorithm}
\end{figure}

Figure \ref{fig:llm_guided_algorithm} shows the overall flow and prompts provided as input to LLM.
%The Guided LLM algorithm is provided with the unstructured CV and JDs.
A configuration is provided to ``guide" the LLM according in the desired objective direction, \textit{``criteria"}. This information together with appropriate prompts is provided as an input to the LLM. 
The output from the LLM is consistent due to the guidelines from the prompts and consists of recommended Job ID along with explanations such as benefits, drawbacks, and qualitative aspects.

\subsection{LLM Unguided Algorithm}
To leverage LLMs ability to comprehend natural language, we perform an experiment where no clear criteria is provided for recommending a job to a talent viz. an ``unguided" approach. This way, the LLM model is expected to recommend jobs that are ``good” according to its own comprehension.
%There is no definite structure to its output, nor it is directed to return a specific number of recommendations. 
%In order to understand the basis on which the model generates recommendations, 
We direct the model in the prompt to add an explanations for the recommendation. With this approach, the model provides recommendations with logical explanations.

%The prompts for this approach are shown in Figure \ref{fig:llm_unguided_algorithm}.

\begin{figure}
\centering
\includegraphics[width = 3in, height = 1.5in]{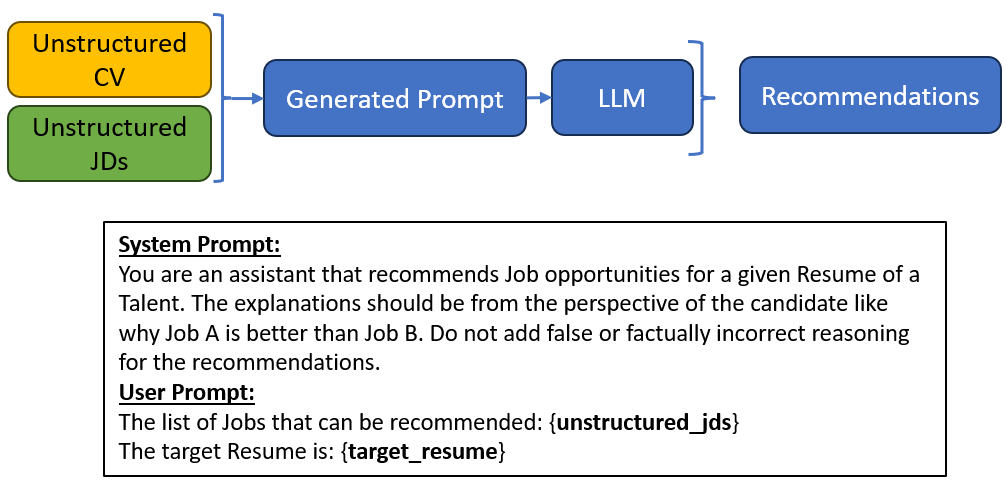}
\caption{Language model based unguided algorithm.}
\label{fig:llm_unguided_algorithm}
\end{figure}

Figure \ref{fig:llm_unguided_algorithm} shows the overall flow and prompts used in this approach.
%Similar to the guided approach, the unstructured CV and unstructured JDs are provided as input, with the key difference being that,
This time no \textit{``criteria"} is provided to guide the model.
The output obtained from this approach does not follow any specific structure. Nevertheless,
%the jobs recommended can easily be identified as
it provides a structure to its response in multiple paragraphs with the Job ID at top and the explanations following below as a bulleted list.

\subsection{Handling large data}
LLMs have a limitation on number of tokens that can be provided as an input. It is 8192 for the OpenAI GPT4. When working with a large number of unstructured JDs, it is inevitable that we will hit this token limit.
%To overcome this issue, following mechanism can be used to handle large number of JDs as an input: 
In that case, we propose to:
%If token limit is exceeded:
(i) Split the JD set to smaller subsets
(ii) Get recommendations separately from each subset
(iii) Merge the top recommendations from each subset.
Note that this merging only top recommendations from each set will overlook superior but lower ranked recommendations from other sets.
%%%%%%%%%%%%%%%%%%%%%%%%%%%%%%%%%%%%%%%%%%%%%%%%%%%%%%%%%%
\section{Hybrid approach}
%%%%%%%%%%%%%%%%%%%%%%%%%%%%%%%%%%%%%%%%%%%%%%%%%%%%%%%%%%
The key idea here is to complement weakness of one technique by the strength of the other.
\begin{itemize}
\item \emph{Qualitative aspects}: Traditional technique is effective in matching well-structured quantitative attributes such as skill proficiency, role, experience, among others. However, they lack in capturing qualitative aspects of talent and job.
Whereas, LLM can comprehend language and take into consideration qualitative attributes and soft skills that are important for a job but that have been missed in the structured model.
Further, they can provide justifications of recommendations in natural language format.
\item \emph{Unbiased view of job and organization}: 
LLM can provide an unbiased view of a job and the organization based on its knowledge gained from training on Internet data.
This can guide the user to take an unbiased informed decision.
\item \emph{Scalability and cost}: LLM based technique faces a drawbacks that it is costly and has scalability challenge due to limitations on number of tokens.
Whereas, the traditional technique is computationally efficient and fast, hence has the capability to process large amount of data.
\end{itemize}

\begin{figure}
\centering
\includegraphics[width = 3in, height = 1.7in]
{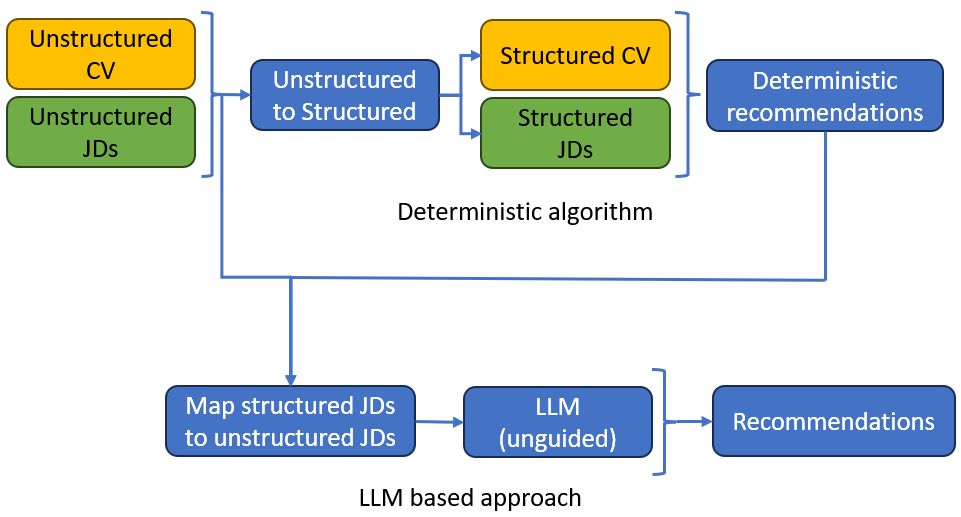}
\caption{Hybrid Algorithm.}
\label{fig:hybrid_algorithm}
\end{figure}

We propose a hybrid approach that combines traditional method with use of language models to generate richer job recommendations.
The key-idea is (i) to use traditional method to trim down large number of job opportunities to a smaller relevant set and (ii) prioritize these further on the basis of qualitative aspects, with well justified reasons, benefits, and drawbacks in natural language.
Figure \ref{fig:hybrid_algorithm} outlines the hybrid approach.
%For the Hybrid algorithm, the inputs remain same as in all other methods consisting of an unstructured CV and unstructured JDs.
Deterministic algorithm is used in the first stage and the unguided LLM method is used in the second stage.
%Figure \ref{fig:hybrid_algorithm} outlines the hybrid approach.

%\begin{enumerate}
%    \item The deterministic algorithm selects a smaller set of JDs.
%    \item The corresponding unstructured JDs for the selected ones are provided as an input to the unguided LLM algorithm.
%    \item LLM unguided approach then selects a subset of job recommendations.
%\end{enumerate}

\begin{figure} [!ht]
\centering
\includegraphics[width = 3in, height = 2in]
{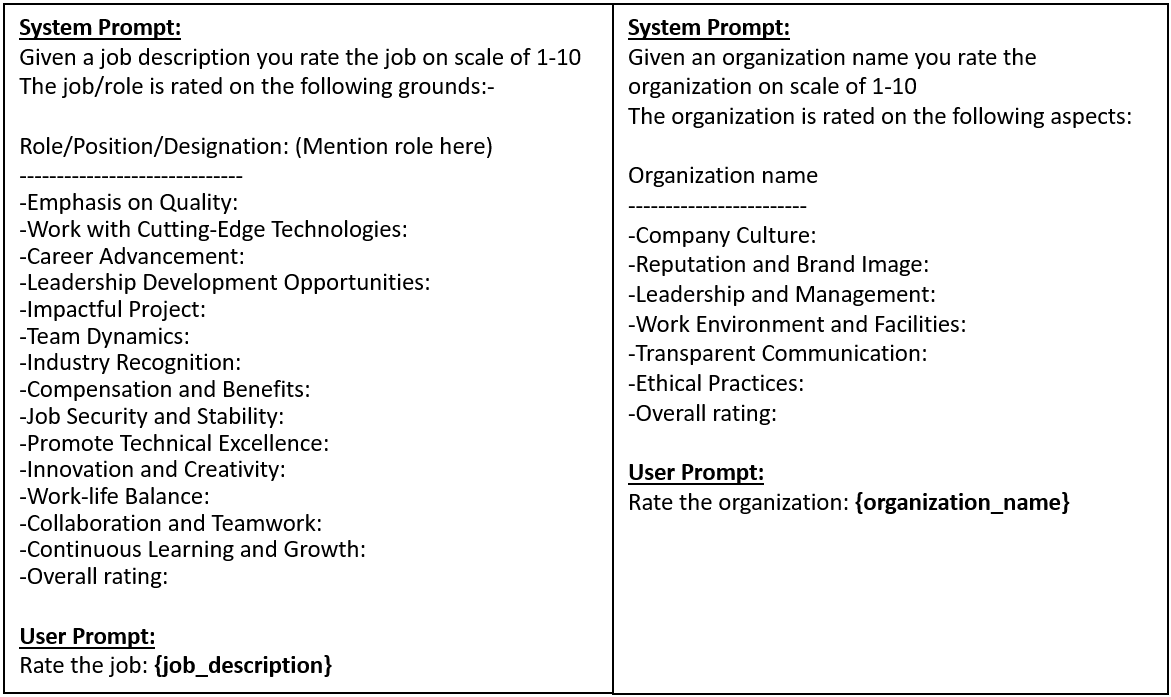}
\caption{Prompts used to generate job and organization ratings}
\label{fig:org_job_rating_prompt}
\end{figure}

The recommendations are further enriched by
an unbiased view of the role and organization
to overcome the issue of
JDs of organizations highlighting only the perks of working for them, without mentioning drawbacks. 
An unbiased view makes it easier for the candidate to make an informed decision. We prompt the LLM to rate the organizations and the particular role in that organization on a scale of 1 to 10. The prompt containing the aspects of the organization and roles that are considered for the rating are shown in Figure \ref{fig:org_job_rating_prompt}.

%In order to make it easier for the candidate to make an informed decision, we prompt the LLM to rate the organizations and the particular role in that organization on a scale of 1 to 10, to have an unbiased view of the role and organization as a whole. 

We derive recommendations on two sets of data viz.
(i) controlled experiments using synthetically generated data
(ii) experiments using real world JDs.
We present the results in upcoming sections.

%%%%%%%%%%%%%%%%%%%%%%%%%%%%%%%%%%%%%%%%%%%%%%%%%%%%%%%%%%
\section{Experiments and Results: Synthetic data}
%%%%%%%%%%%%%%%%%%%%%%%%%%%%%%%%%%%%%%%%%%%%%%%%%%%%%%%%%%
\subsection{Data generation} \label{subsec:synthetic_data_generation}

%\begin{figure}
%\centering
%\includegraphics[width = 2.5in, height = 1.4in]{./figures/synthetic_data.png}
%\caption{Synthetic data generation.}
%\label{fig:synthetic_data}
%\end{figure}

\begin{figure}
\centering
\includegraphics[width = 3.3in, height = 1.6in]
{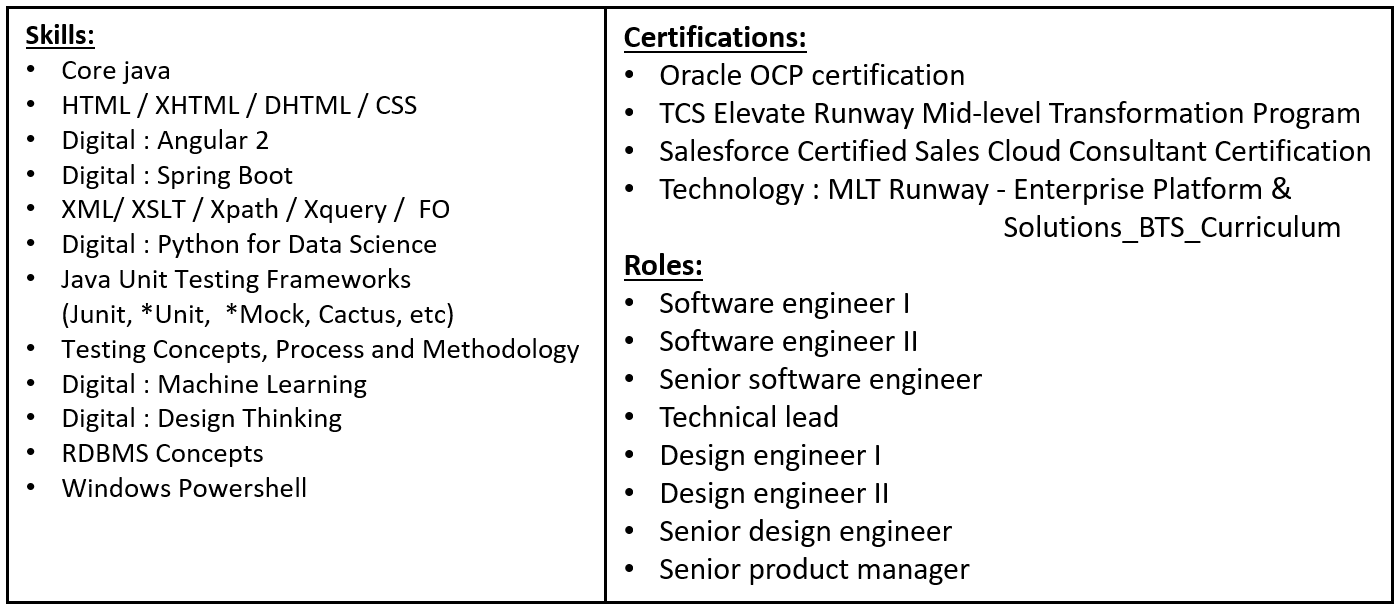}
\caption{Set from which attribute values are sampled.}
\label{fig:syn_data_attributes}
\end{figure}

\begin{figure}
\centering
\includegraphics[width = 3.5in, height = 2.5in]
{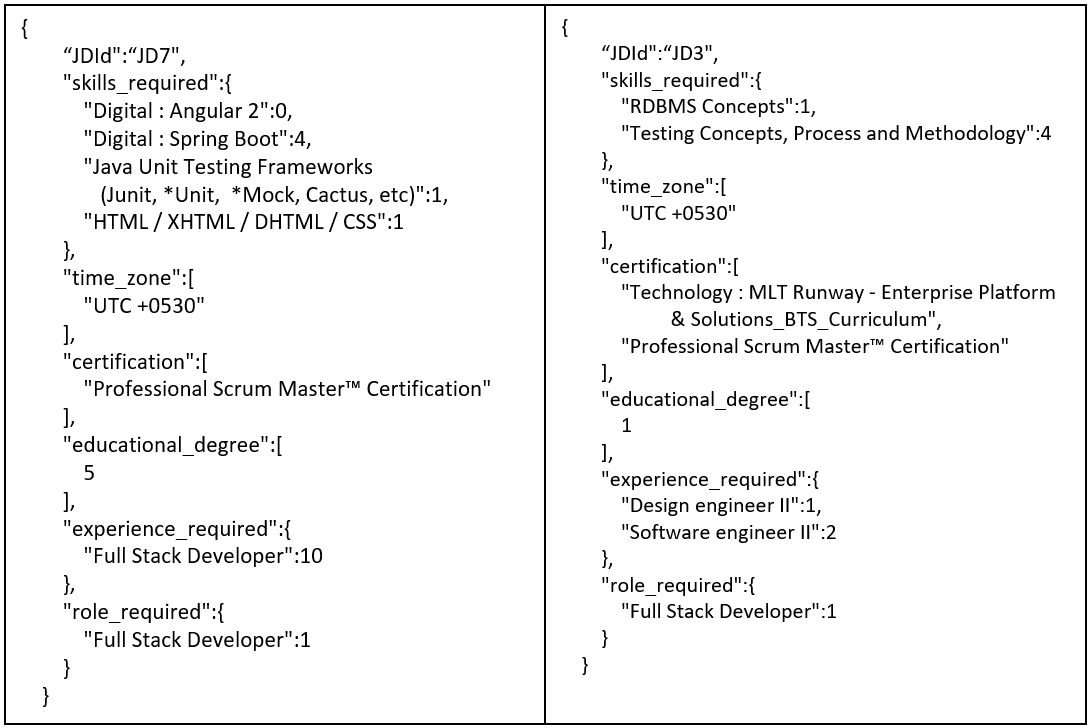}
\caption{Structured synthetic JDs: JD7 and JD3.}
\label{fig:structured_five_jds_for_cv3}
\end{figure}

%Figure \ref{fig:synthetic_data} describes the synthetic data generation process.
We model the domain of Information Technology by populating relevant values of attributes. The list of values and their ranges are shown in Figure \ref{fig:syn_data_attributes}.
We sample from these values to populate structured models of talent and job (Figure \ref{fig:structured_resume_for_cv3}, \ref{fig:structured_five_jds_for_cv3}).

For the purpose of performing controlled experiments, we generate ten JDs for every CV.
%and derive five job recommendations for each of them. 
The jobs are generated in a way such that, one JD is almost an exact match, three JDs deviate a little, the next three have larger deviation, the last three are quite different.
The traditional deterministic algorithm is expected to recommend and rank the JDs in above order..

\begin{figure}
\centering
\includegraphics[width = 3.5in, height = 2in]
{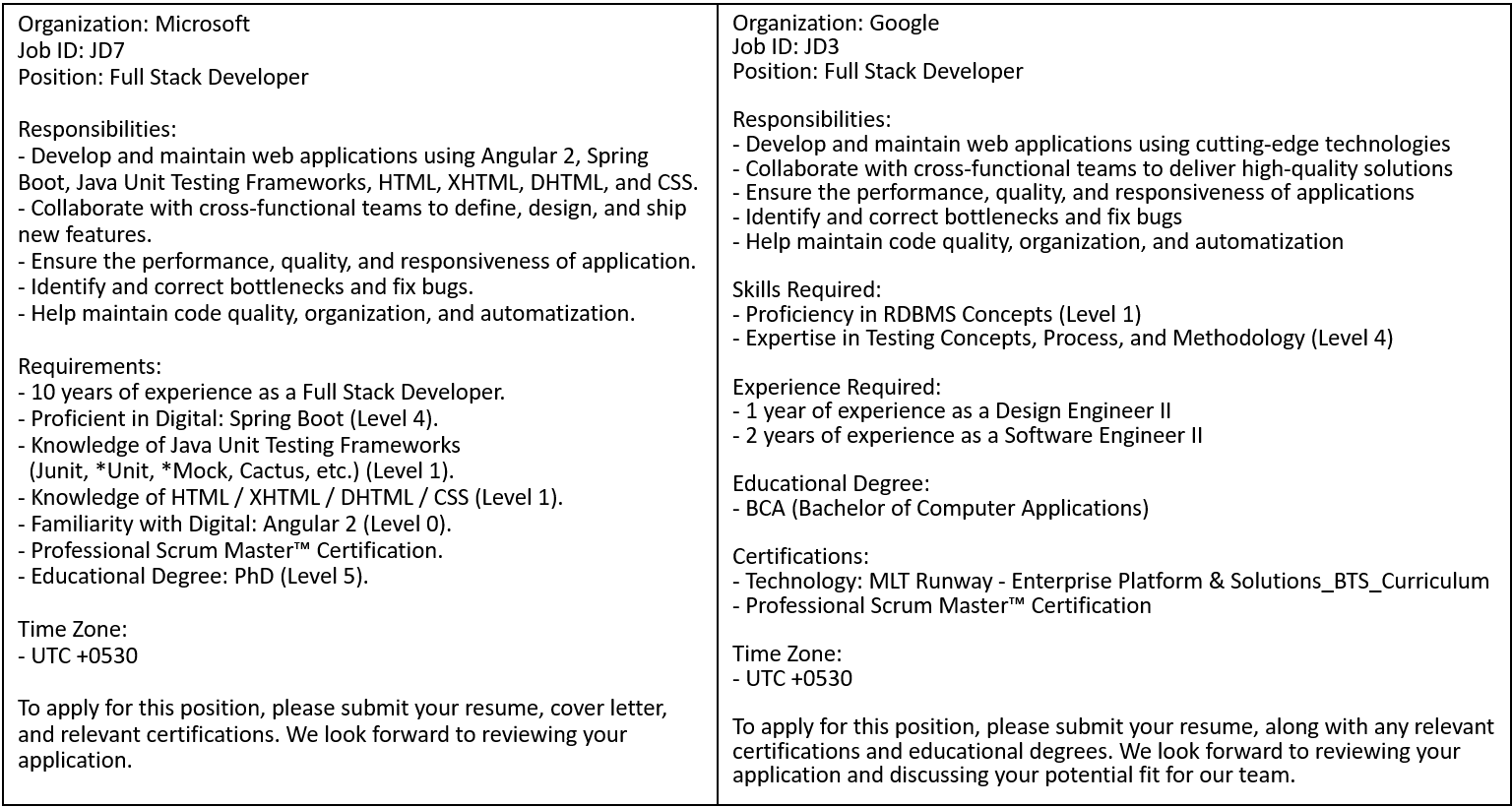}
\caption{Unstructured synthetic JDs: JD7 and JD3.}
\label{fig:unstructured_five_jds_for_cv3}
\end{figure}

\begin{figure}
\centering
\includegraphics[width = 3in, height = 2.5in]
{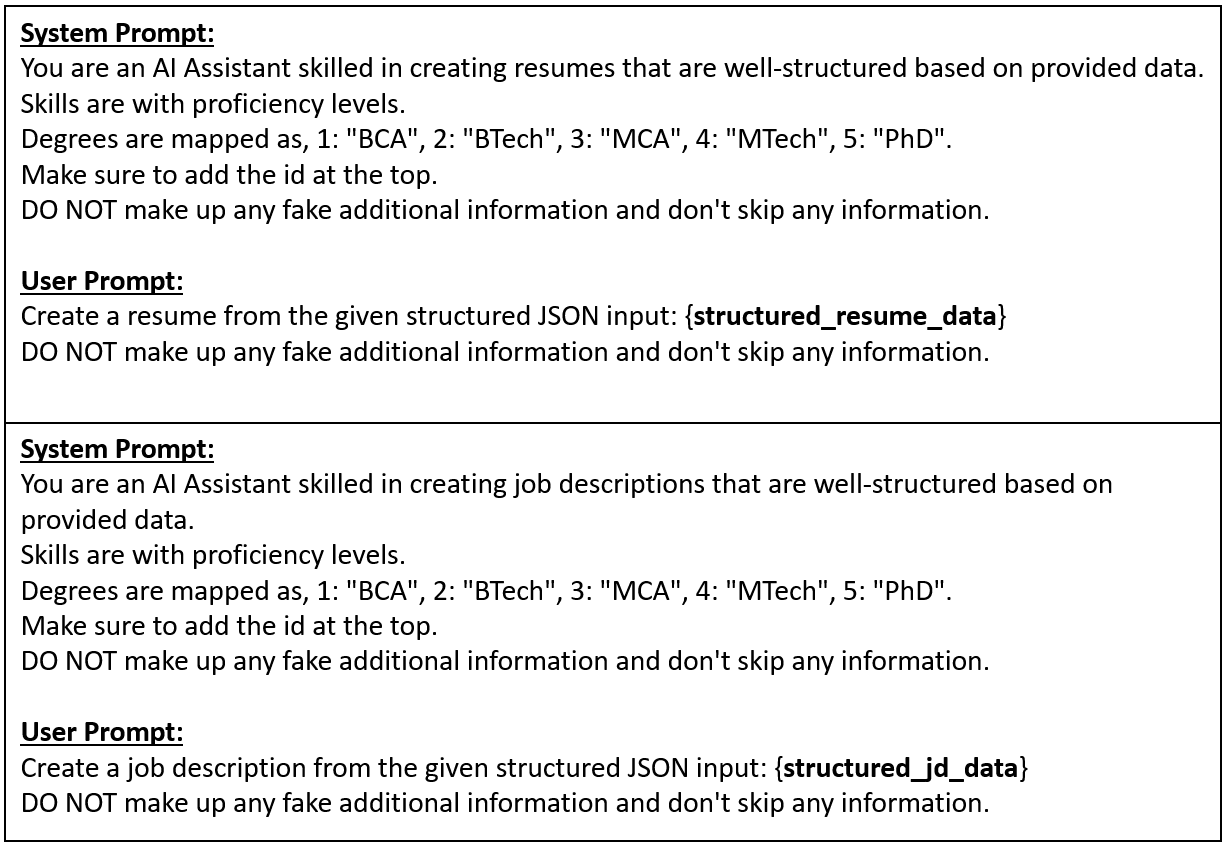}
\caption{Synthetic data generation prompts for creating unstructured CV (top) and unstructured JD (bottom).}
\label{fig:synthetic_data_generation_prompt}
\end{figure}

We generated realistic unstructured CV and JDs from the structured talent and JD models using GPT4. These provided as input to LLMs.
%, as this better represents real world data.
The prompts used for this purpose is shown in Figure \ref{fig:synthetic_data_generation_prompt}.
Figure \ref{fig:unstructured_five_jds_for_cv3} shows unstructured JDs corresponding to structured JDs in Figure \ref{fig:structured_five_jds_for_cv3}.

\subsection{Results}
\begin{figure}
\centering
\includegraphics[width = 1.5in, height = 0.9in]
{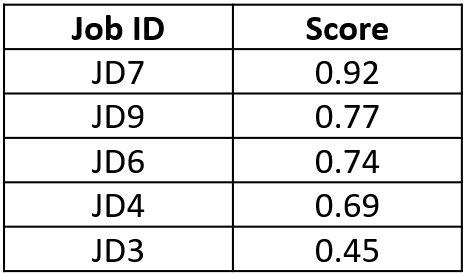}
\caption{Deterministic algorithm recommendations for synthetic CV3.}
\label{fig:deterministic_algo_recos_for_cv3}
\end{figure}

We present details of recommendations generated by each algorithm for one candidate (CV3).
Refer Figure \ref{fig:deterministic_algo_recos_for_cv3} for recommendations by deterministic algorithm.
JD7 is a perfect match for CV3, with all attributes matching and just slight variation in the skill proficiencies.
JD9 is the next best match, requiring better skill proficiency, and more certifications.

\begin{figure}
\centering
\includegraphics[width = 3.5in, height = 2.5in]
{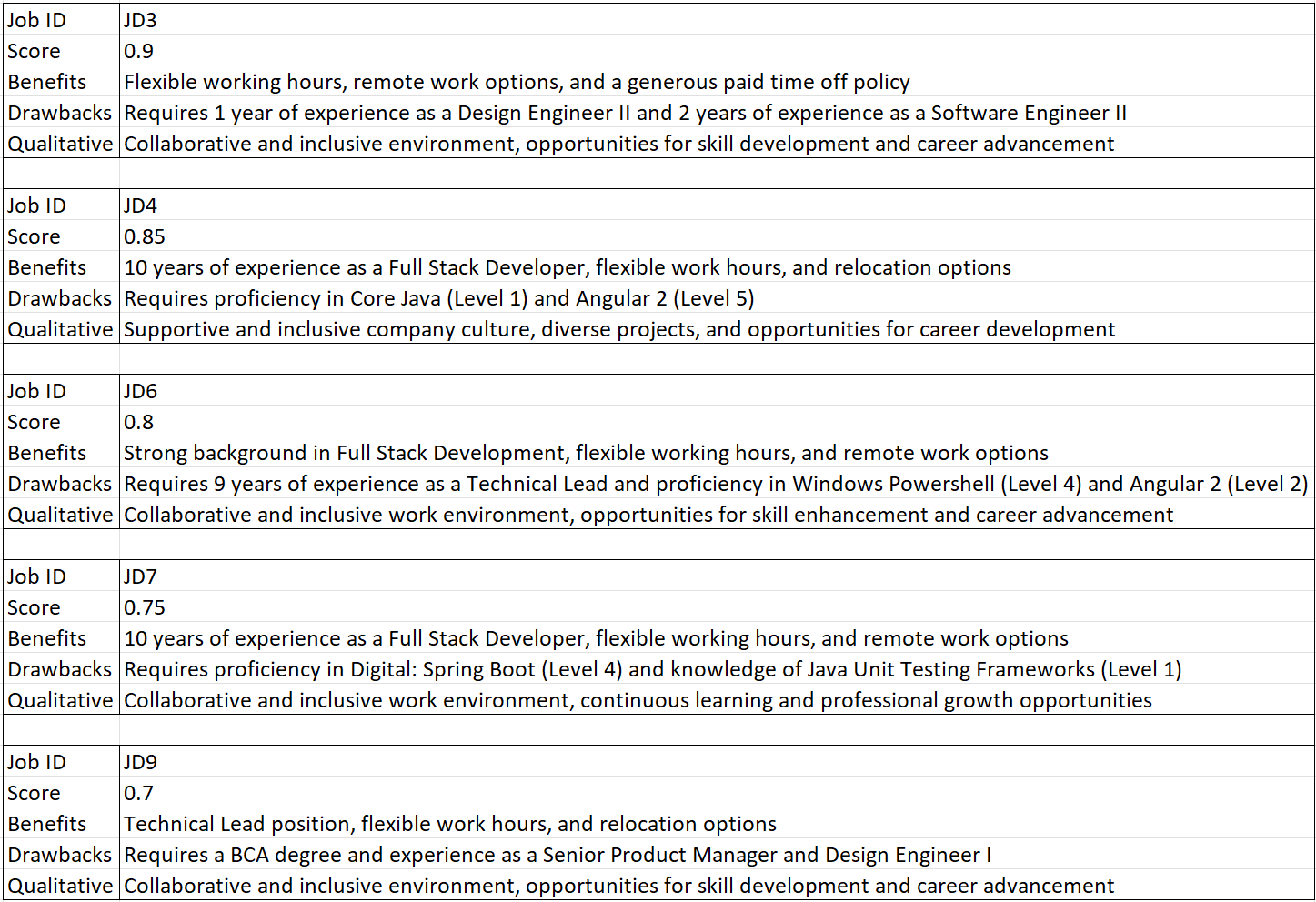}
\caption{LLM guided recommendations for synthetic CV3.}
\label{fig:llm_guided_recos_for_cv3}
\end{figure}

The language model pays more attention to the role match, experience and education.
Refer Figure \ref{fig:llm_guided_recos_for_cv3} for recommendations by ``guided" approach.
It has recommended JD3 which isn't a good match by deterministic algorithm, with only the role requirement and timezone conditions being satisfied.
However, it has weighted the qualitative attributes that are not captured in the structured format viz.
collaborative and inclusive environment, opportunities for skill development and career advancement.
One aspect it seems to miss out across all its recommendations is the mismatch in representation of proficiency level.
It is unable to understand the terminology of beginner, intermediate, advanced level and compare it with numeric proficiency levels between 0 and 5.
This behaviour is expected as domain dependent numeric values do not make sense unless the range of the values or its mapping with textual classifications are mentioned in advance.
It has recommended all the JDs that have preferred role by the candidate with CV3 viz. Full Stack Developer and Technical Lead.

\begin{figure}
\centering
\includegraphics[width = 3.5in, height = 2.2in]
{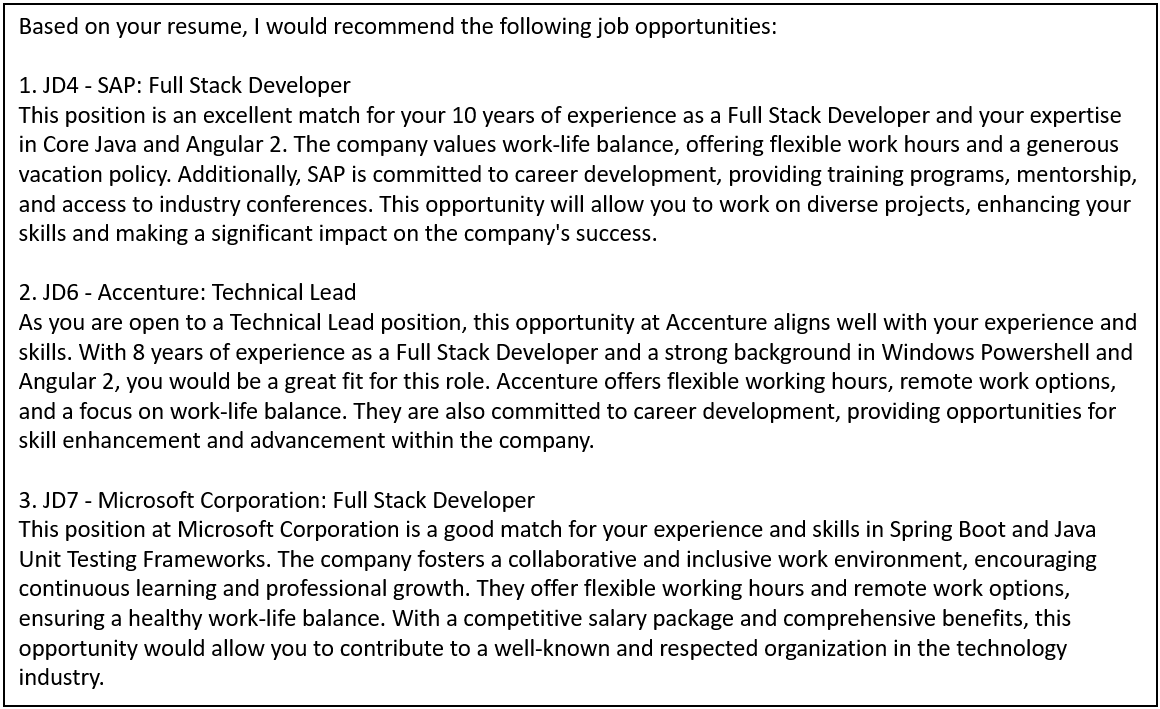}
\caption{LLM unguided recommendations for synthetic CV3.}
\label{fig:llm_unguided_recos_for_cv3}
\end{figure}

Refer Figure \ref{fig:llm_unguided_recos_for_cv3} for recommendations by ``unguided" approach.
In this case, JD4, JD6, JD7 have been recommended and the recommendations are well justified.
Again, it has given more preference to role, and experience.
The qualitative aspects of organizations is correctly captured and presented in greater detail.
Being completely unstructured, it clearly explains the perks of the recommended jobs and also mentioned the reasons why the others were not a good match.
%Finally, not everytime though, it ends off the response with a good luck note for a successful job hunt.

\begin{figure}
\centering
\includegraphics[width = 3.5in, height = 2.9in]
{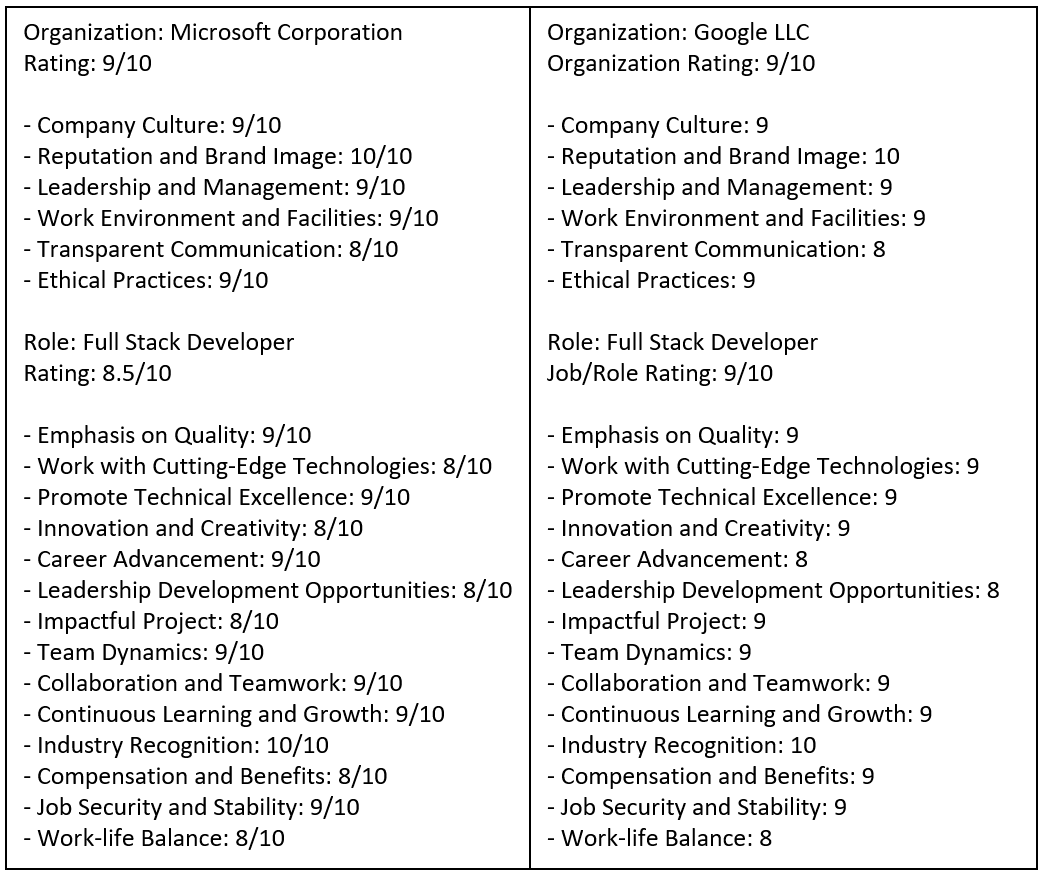}
\caption{Organization and job ratings in hybrid algorithm recommendations for synthetic CV3.}
\label{fig:org_job_rating_hybrid}
\end{figure}

Figure \ref{fig:llm_unguided_recos_for_cv3} presents recommendations generated by the hybrid algorithm.
The deterministic algorithm recommends 5 JDs. The LLM unguided method then picks out the best 3 from among those. 
In addition to this, the algorithm also provides ratings of qualitative aspects of organizations and job traits as shown in Figure \ref{fig:org_job_rating_hybrid}.
%, by prompting the LLM seperately with the prompts in Figure \ref{fig:org_job_rating_prompt}.
Google and its Full Stack Developer position is rated highest 9 on 10 while SAP has organization rating of 8.7 and
its Full Stack Developer position has a rating of 8.6. Finally, TCS has organization rating of 8.5 but the position of Technical lead is rated 8.7.
This helps the candidate have an unbiased view and take an informed decision. 

\begin{figure}
\centering
\includegraphics[width = 3.5in, height = 2.5in]
{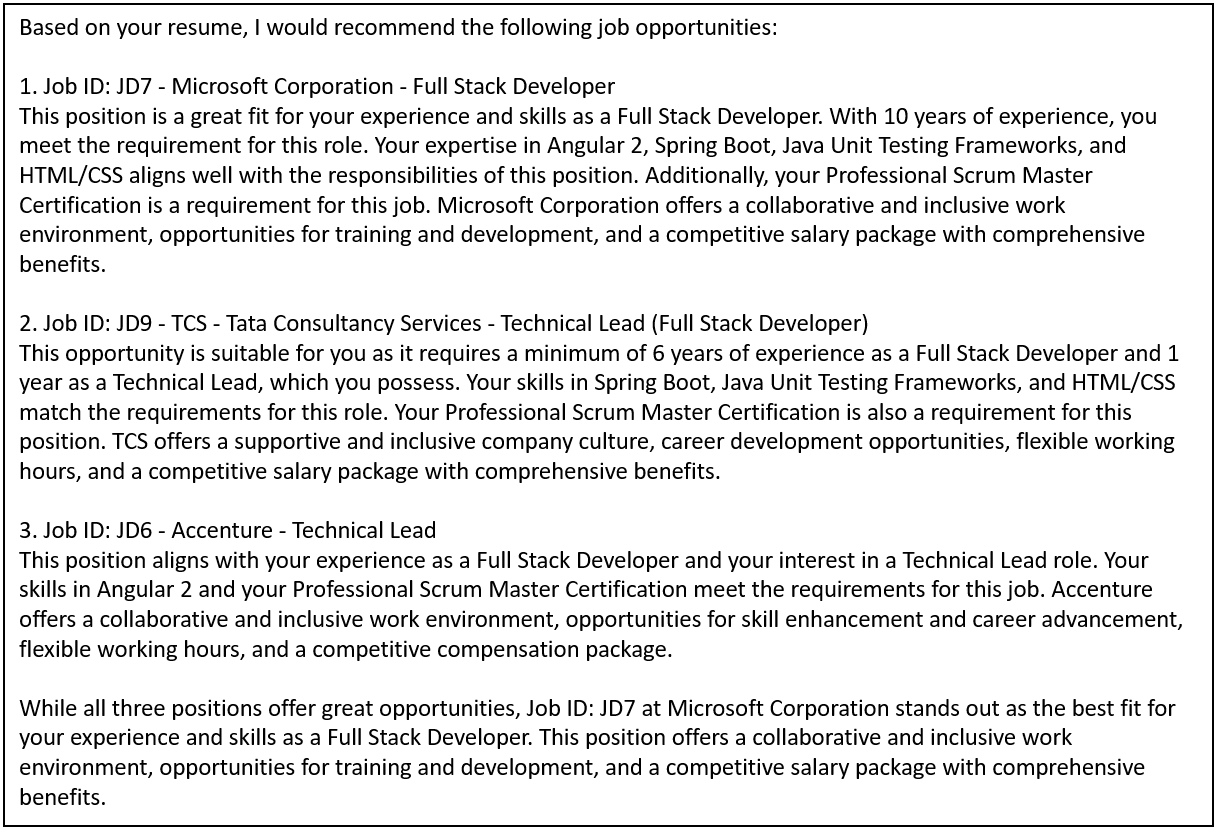}
\caption{Hybrid recommendations for synthetic CV3.}
\label{fig:hybrid_recos_for_cv3}
\end{figure}

%%%%%%%%%%%%%%%%%%%%%%%%%%%%%%%%%%%%%%%%%%%%%%%%%%%%%%%%%%
\section{Real world experiments}
%%%%%%%%%%%%%%%%%%%%%%%%%%%%%%%%%%%%%%%%%%%%%%%%%%%%%%%%%%
\subsection{Data description}
\begin{figure}
    \centering
    \includegraphics[width = 3.2in, height = 1.2in]{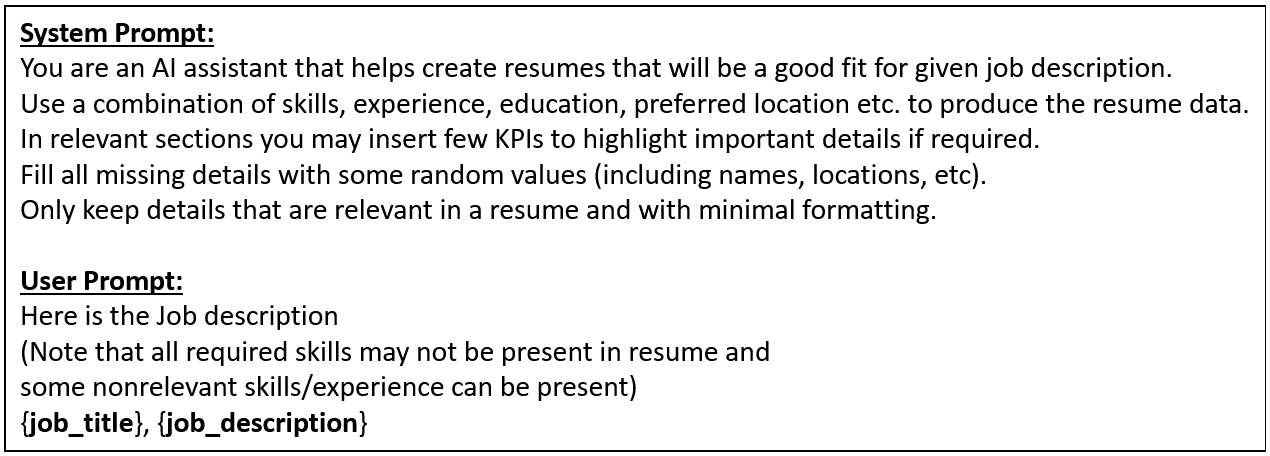}
    \caption{Prompt for unstructured CV generation from a real unstructured JD.}
    \label{fig:unstructured_real_cv_gen_prompt}
\end{figure}

We used real JDs from Kaggle \cite{kaggle} and
generated CVs from these.
%From the multiple jobs present in the dataset first 10 were picked for this experiment.
%Refer Figure \ref{fig:real_data_generation} for the data generation process.
The unstructured CVs are generated using LLM. The prompt used for this is presented in Figure \ref{fig:unstructured_real_cv_gen_prompt}.
The JDs have a lot of variation, which helps capture real world situations where JDs can be completely different.

%\begin{figure} [!ht]
%\centering
%\includegraphics[width = 2.8in, height = 1.3in]{./figures/real_data_generation.png}
%\caption{Real data generation.}
%\label{fig:real_data_generation}
%\end{figure}

We used five CVs to generate recommendations.
For three of the CVs, we did not include the source JDs from which they are generated, whereas for two CVs, we included the source JDs.
The thought behind this was to have some CVs which would not have a one to one match with any of the available JDs and the recommenders would be forced to pick some next best option. This would help make this experiment more relevant and better suited to capture real world scenario.

\subsection{Results}
\begin{figure}
\centering
\includegraphics[width = 3.4in, height = 3.2in]
{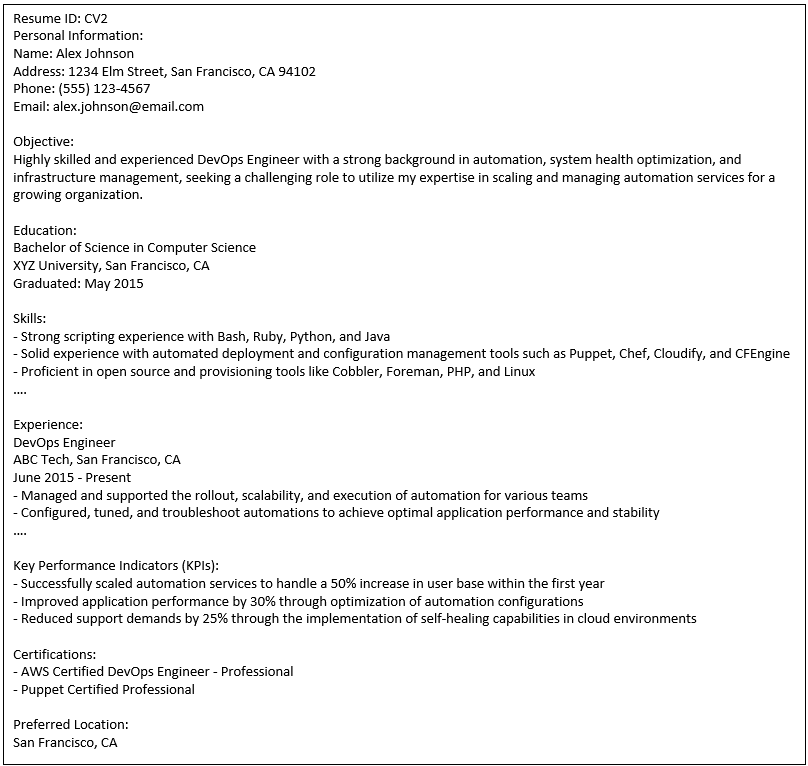}
\caption{Unstructured real CV2.}
\label{fig:unstructured_real_cv2}
\end{figure}

\begin{figure} [h]
\centering
\includegraphics[width = 2.7in, height = 2.9in]
{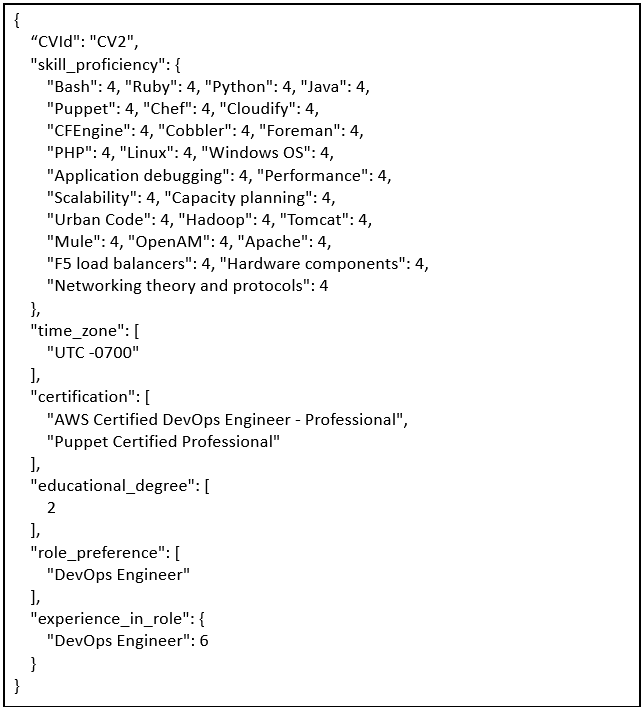}
\caption{Structured real CV2.}
\label{fig:structured_real_cv2}
\end{figure}

\begin{figure}
\centering
\includegraphics[width = 3.4in, height = 4.6in]
{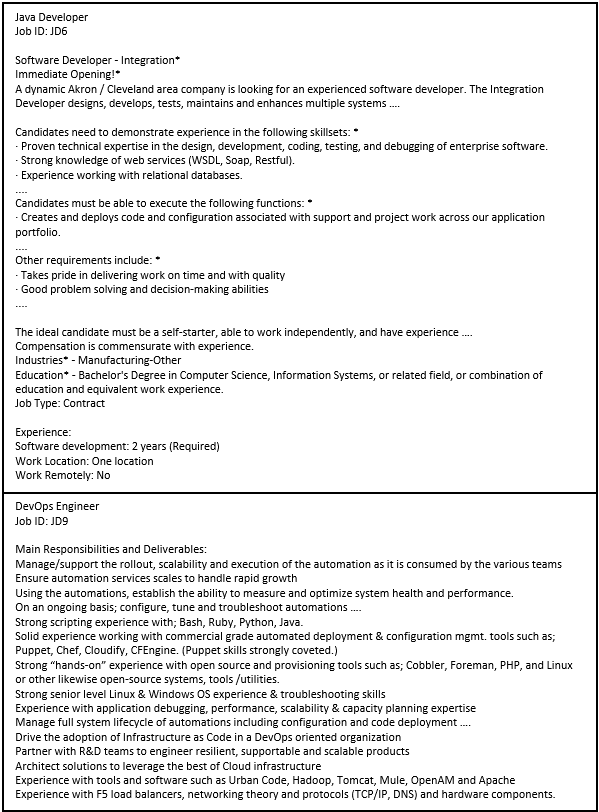}
\caption{Unstructured real JD6 and JD9.}
\label{fig:unstructured_real_five_jds}
\end{figure}

\begin{figure} [h]
\centering
\includegraphics[width = 3.3in, height = 2.5in]
{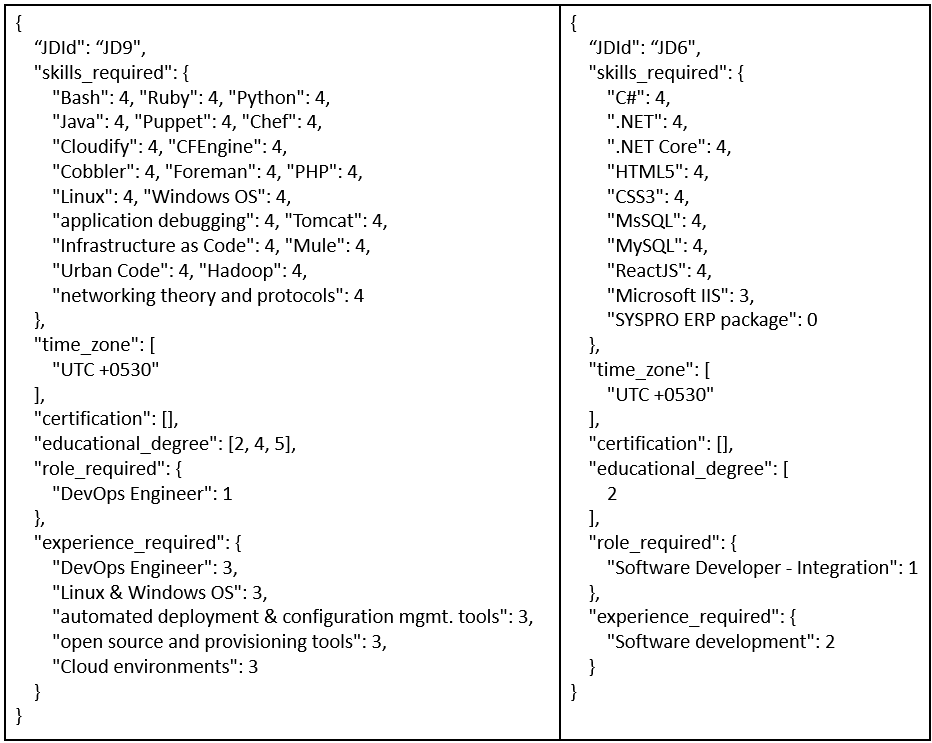}
\caption{Structured real JD9 and JD6.}
\label{fig:structured_real_five_jds}
\end{figure}

\begin{figure}
\centering
\includegraphics[width = 1.5in, height = 0.9in]
{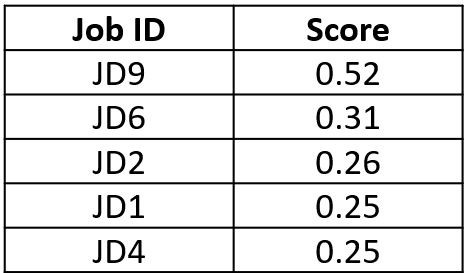}
\caption{Deterministic algorithm recommendations for CV2.}
\label{fig:deterministic_results_real_cv2}
\end{figure}

\begin{figure}
\centering
\includegraphics[width = 3.5in, height = 2.7in]
{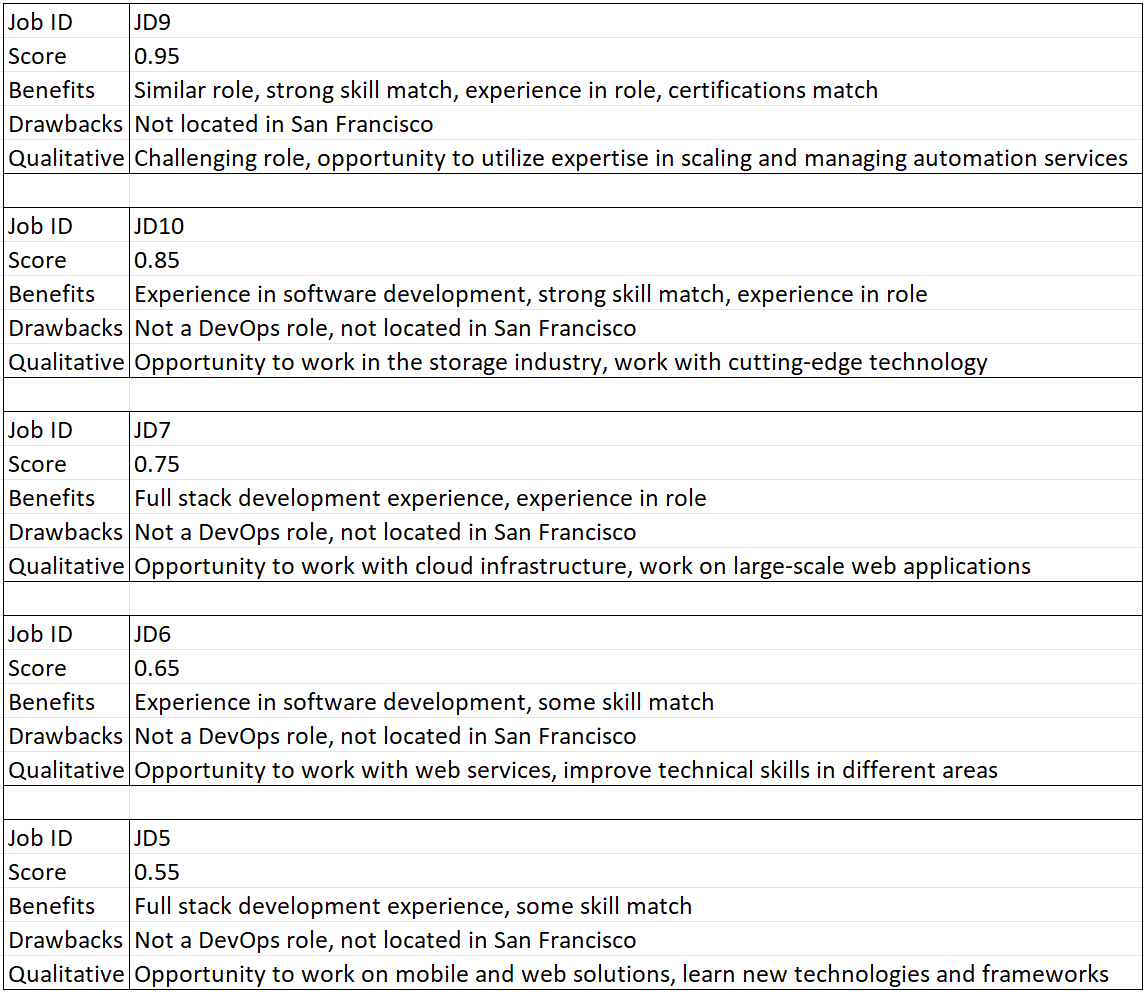}
\caption{LLM guided algorithm recommendations for real CV2.}
\label{fig:llm_guided_results_real_cv2}
\end{figure}

\begin{figure}
\centering
\includegraphics[width = 3.5in, height = 2.4in]
{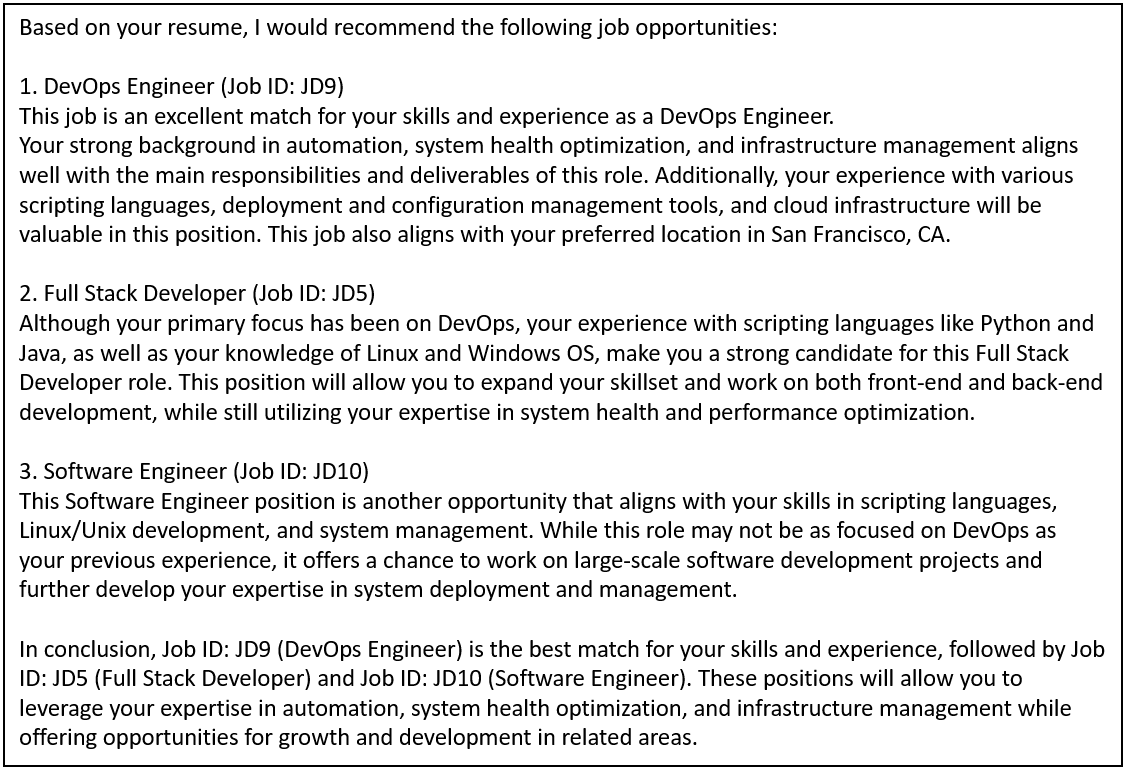}
\caption{LLM unguided algorithm recommendations for real CV2.}
\label{fig:llm_unguided_results_real_cv2}
\end{figure}

\begin{figure}
\centering
\includegraphics[width = 3.5in, height = 2.4in]
{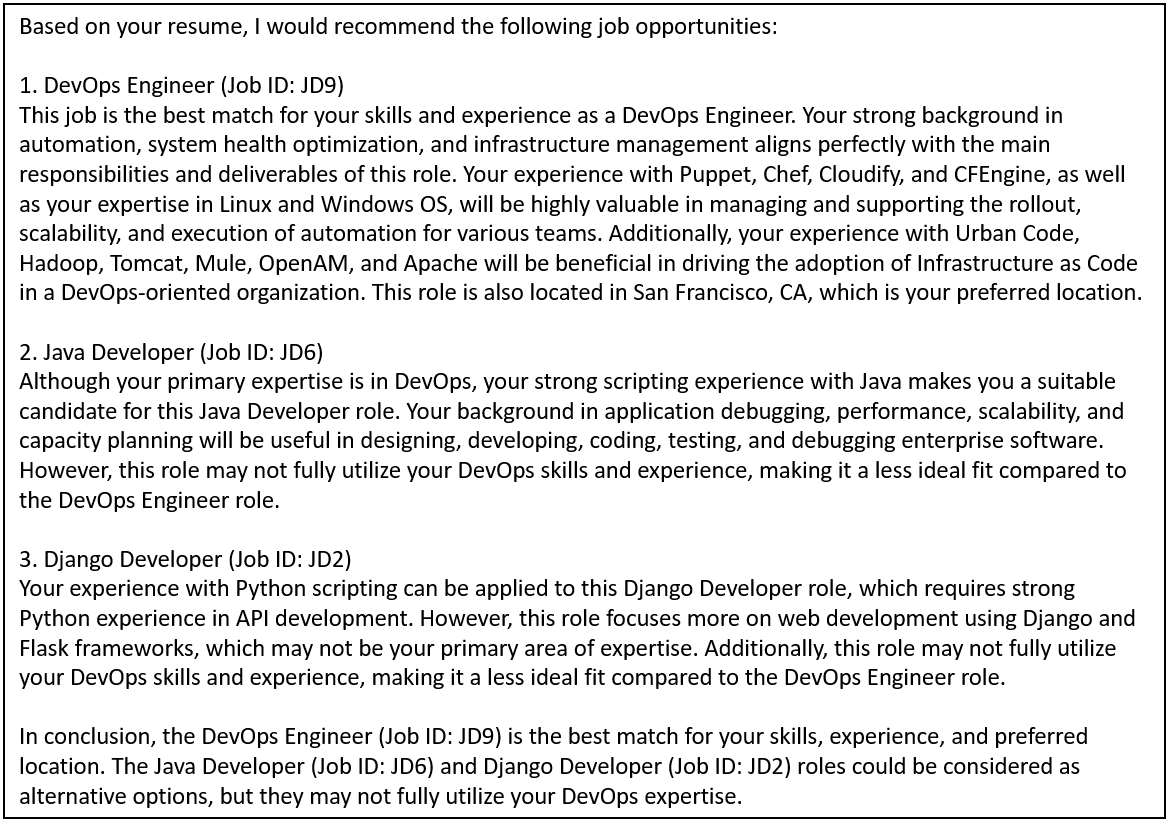}
\caption{Hybrid approach recommendations for real CV2.}
\label{fig:hybrid_results_real_cv2}
\end{figure}

We present details of recommendations generated by each algorithm for one CV (CV2) shown in Figure \ref{fig:unstructured_real_cv2} and \ref{fig:structured_real_cv2}.
%for the unstructured and corresponding structured CV of a candidate.
Figure \ref{fig:unstructured_real_five_jds} show two unstructured JDs and Figure \ref{fig:structured_real_five_jds} shows corresponding structured JDs.
Figure \ref{fig:deterministic_results_real_cv2},
\ref{fig:llm_guided_results_real_cv2},
\ref{fig:llm_unguided_results_real_cv2}, and
\ref{fig:hybrid_results_real_cv2} show the recommendations by deterministic, LLM-guided, LLM-unguided, and Hybrid approach respectively.

The deterministic and LLM approaches recommend JD9 from which CV2 was generated. It is best with respect to skill match and role of ``DevOps Engineer".
In was observed that none of the JDs had any specific location requirement.
As a result, LLM took liberty to show it as either a drawback or a benefit. In the LLM-guided approach, the method brings it out as a drawback. However, the LLM-unguided approach and Hybrid approach bring it out as a benefit.
LLM-guided approach points out that the job opportunity is not at the preferred location ``San Francisco".
However, LLM-unguided approach and Hybrid approach mention that there a match in preferred location.
Hybrid algorithm brings out the positives of JD9 with respect to the skill match and how particular skills such as
Hadoop, Urban code, Tomcat will help in ``Infrastructure as Code in a DevOps oriented organization". 
Rest of the JDs recommended are not as good a match compared to JD9.
However, JD6 has been recommended on the basis of similarity of skills of DevOps Engineer and Java Developer or Software Developer position.

\begin{figure}
\centering
\includegraphics[width = 3.4in, height = 1.5in]
{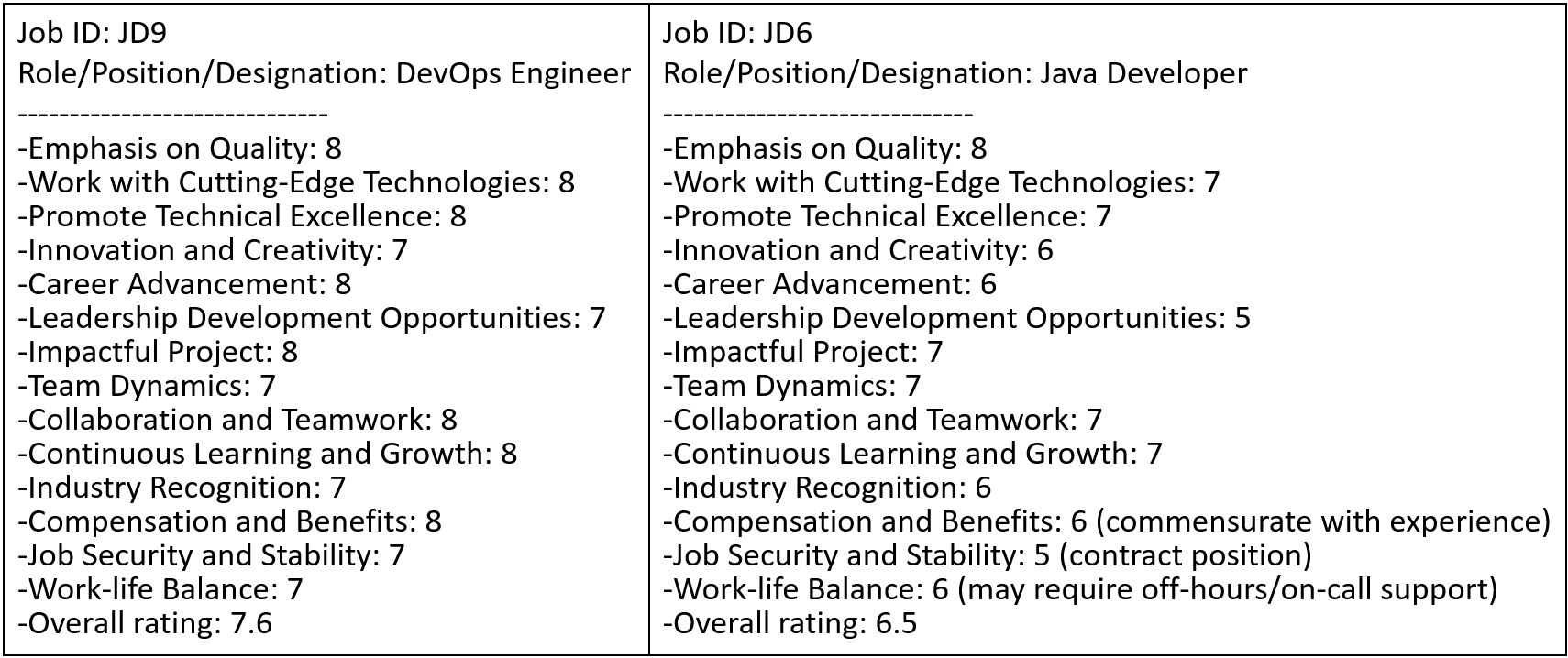}
\caption{Job ratings for positions recommended for real CV2.}
\label{fig:job_ratings_real_data}
\end{figure}

Figure \ref{fig:job_ratings_real_data} shows the job ratings produced by LLM in Hybrid algorithm.
Clearly, the position of ``DevOps Engineer" in JD9 has a higher rating of 7.6 compared to a ``Java Developer" position in JD6.

%%%%%%%%%%%%%%%%%%%%%%%%%%%%%%%%%%%%%%%%%%%%%%%%%%%%%%%%%%
\section{Ballpark estimates of quality of algorithm}
%%%%%%%%%%%%%%%%%%%%%%%%%%%%%%%%%%%%%%%%%%%%%%%%%%%%%%%%%%
\label{sec:ballpark_estimate_calculation}
Assigning a quality score to a recommendation without human feedback is hard. We assign reference quality scores manually to each recommendation.
The manual reference scores are assigned based on how a human would have rated the jobs for a given CV, i.e., by considering all ``must-have” conditions first, like skills, education, experience and then focusing on the ``good-to-have” like location, certification and other associated benefits provided in the job package. However, since these scores are manually assigned there can be oversights along with aspects of jobs that can be interpreted differently depending on the individual rating the jobs. So, rather than treating the scores as ground truth, we use them as guidelines and assume that some deviation from them is acceptable.

The quality scores of algorithms (Figure \ref{fig:result_stats_synthetic_data}
and Figure \ref{fig:result_stats_synthetic_data}) are computed
%be comparing the scores generated by algorithm with manual reference scores.
%The quality score is computed
as an average of absolute deviation of algorithm scores from the manual reference scores.
In case of LLM unguided algorithm that just provide rank of recommendation the algorithm score is assigned a value same as the manual reference score for the same rank.

\subsection{Synthetic data}
\begin{figure} [!ht]
\centering
\includegraphics[width = 3.2in, height = 0.8in]{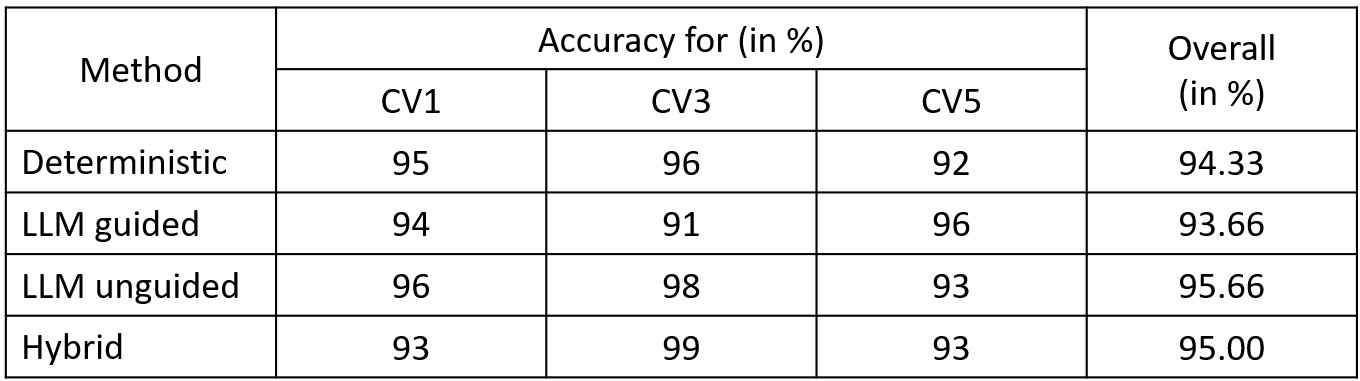}
\caption{Quality estimate of recommendations on synthetic data.}
\label{fig:result_stats_synthetic_data}
\end{figure}

%We did random sampling of results to validate overall quality.
Figure \ref{fig:result_stats_synthetic_data} shows our findings.
%For deatils on how the accuracies are calculated, please refer to section \ref{sec:ballpark_estimate_calculation}.
The results from the four methods exhibit striking similarities. Among them, the LLM unguided approach stands out for its superior performance, closely resembling human-friendly recommendations, and having a accuracy of 95.66\%. We observed that LLMs excel in utilizing domain knowledge and capturing unstructured elements, showcasing their greatest strengths. Conversely, it is evident that constraining the LLM in guided approach yields the poorest results among the four methods, with around 93.66\% accuracy. Deterministic method is better suited for large scale, rendering the use of LLMs inefficient and unjustifiable in terms of time and cost. Consequently, Hybrid approach, leverages the unguided LLM method in conjunction with deterministic methods to combine the best aspects of both approaches, generating results almost identical to the best, with an accuracy of 95\% which is just shy of 0.66\% from the best.

\subsection{Real world data}
\begin{figure}
\centering
\includegraphics[width = 3.2in, height = 0.8in]{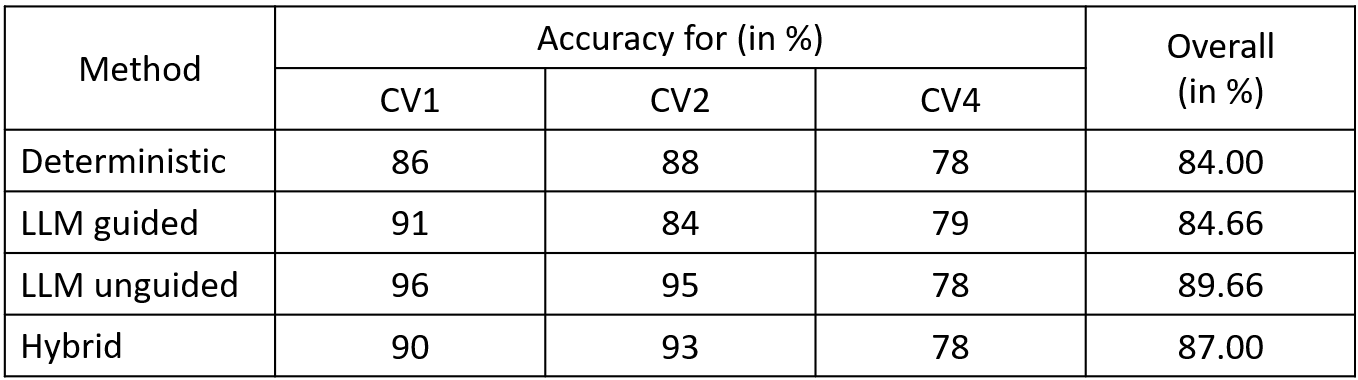}
\caption{Quality of language based recommendations on real data.}
\label{fig:result_stats_real_data}
\end{figure}

%Similar to synthetic data experiments, we randomly sampled a few CVs from the real data to validate the overall model performance.
Our findings for recommendations for real world data are depicted in Figure \ref{fig:result_stats_real_data}.
The quality score on real data is somewhat lower compared to the synthetic data.
%Nevertheless, the consistent trends and differences among the methods are more pronounced in these results.
Similar to the synthetic data, LLM unguided once again outperforms all other methods with an impressive accuracy of 89.66\%. The hybrid method secures the second-best position with an overall accuracy of 87\%, following the same trend as before. LLM guided, on the other hand, falls in the lower half with an accuracy of 84.66\%, marginally outperforming the deterministic method which exhibited the poorest performance this time. The subpar results of the deterministic approach highlight the loss of information during the conversion to a structured data format.

An additional interesting observation can be made from these results. There is a noticeable decline in performance for one of the CVs, CV4, primarily due to the absence of an exact job match. Recommendations for this talent had to rely solely on other relevant factors and general knowledge to suggest a suitable job. Interestingly, all the methods performed equally poorly on this task.
%We speculate that the similar performance between deterministic and LLMs stems from the fact that LLMs were used to generate the structured data for the deterministic method, capturing the same details utilized internally by LLMs for recommendations.

In conclusion, it is evident that LLM unguided exhibits the best performance. However, for practical purposes and the ability to handle larger data volumes, the Hybrid algorithm proves to be a more suitable choice.

%%%%%%%%%%%%%%%%%%%%%%%%%%%%%%%%%%%%%%%%%%%%%%%%%%%%%%%%%%
\section{Performance Analysis}
%%%%%%%%%%%%%%%%%%%%%%%%%%%%%%%%%%%%%%%%%%%%%%%%%%%%%%%%%%
We evaluate runtime performance of all the approaches.
We observed a strong correlation between the number of tokens used in the LLMs, both input and output, and the time it takes to generate recommendations.

\begin{itemize}
    \item Deterministic algorithm: In this case, the time taken by the algorithm itself is negligible when compared to the time required to convert unstructured to structured data. On an average around 20 seconds are required to convert a given CV or a JD to structured format.
    However, in real, JDs may not be required to be converted to structured format every single time the recommendation system runs. %When new jobs are added to the pool of jobs they will be converted to structured form and stored. A talent asking for recommendation would only need to wait around 20 seconds for their CV to be converted into structured form and recommendations to appear. Hence, final performance would be slightly better when compared to other algorithms, even when time taken by LLM is considered in the equation.
    \item LLM guided and LLM unguided: Both of these methods exhibit similar time requirements, although LLM unguided takes slightly longer due to its tendency to employ verbose reasoning, resulting in a higher token count during the process. Generating recommendations for a given CV typically takes an average of 25 to 35 seconds using either method. %Since these approaches operate on unstructured data, there is no additional time needed for converting the data into a structured format.
    \item Hybrid: The hybrid method combines both deterministic and LLM approaches, executing the process sequentially, which results in the cumulative time required by both methods. To run the deterministic part and select good recommendations, it takes approximately 20 seconds. %primarily spent on converting the unstructured CV into a structured format. It is assumed that a repository of structured JDs has been built beforehand, as discussed earlier. 
    Subsequently, the LLM part is invoked, which takes an additional 25 seconds, resulting in a total time of approximately 45 seconds for a single recommendation. While this duration may appear large, it should be noted that even with hundreds of jobs to recommend from, the time will remain relatively constant. %The deterministic part will still take 20 seconds to narrow down the set, upon which the LLM approach operates.
\end{itemize}

%Another crucial aspect that necessitates attention is the reliance on response times from OpenAI servers. The runtime of the system can vary significantly from one recommendation to another, depending on the server's workload at the given moment. A noteworthy instance of this variability occurred when obtaining a recommendation for the synthetic CV4 using the guided LLM method. In this case, the runtime escalated from the usual 30 seconds to approximately 1.5 minutes, which is three times the normal duration. To mitigate such fluctuations, investing in dedicated servers and prioritized traffic can contribute to reducing the overall time required for LLMs to generate results.

%%%%%%%%%%%%%%%%%%%%%%%%%%%%%%%%%%%%%
\section{Literature Review}
%%%%%%%%%%%%%%%%%%%%%%%%%%%%%%%%%%%%%

% Content based JRS:
% \begin{itemize}
%     \item Bag of Words: \cite{mpela2020mobile}, \cite{kessler2012hybrid}
%     \item Latent Dirichlet Allocation: \cite{bansal2017topic}
%     \item Word2vec: \cite{gugnani2020implicit}, \cite{valverde2018job}
%     \item Observation: \cite{schmitt2016matching}
% \end{itemize}

% Collaborative filtering JRS:
% \begin{itemize}
%     \item KNN based approach: \cite{lacic2020using}, \cite{lee2017exploiting}, \cite{reusens2017note}
% \end{itemize}

% Hybrid JRS:
% \begin{itemize}
%     \item Model-based methods on shallow embeddings: \cite{chen2017hybrid}, \cite{maheshwary2018matching}, \cite{poch2014ranking}, \cite{yang2017combining}
%     \item Deep neural networks: \cite{jiang2020learning}, \cite{luo2019resumegan}, \cite{nigam2019job}, \cite{qin2018enhancing}, \cite{zhu2018person}
%     \item Ensemble hybrids: \cite{chen2016xgboost}, \cite{de2016scalable}, \cite{lu2013recommender}, \cite{volkovs2017content}, \cite{xiao2016job}
%     \item Weighted hybrids: \cite{leksin2016job}, \cite{leksin2017combination}, \cite{polato2016preliminary}
%     \item Some other: \cite{borisyuk2017lijar}, \cite{gui2016downside}, \cite{laumer2018job}
%  \end{itemize}

% Knowledge-based JRS: \cite{freire2021recruitment}, \cite{kmail2015matchingsem}, \cite{martinez2020novel}, \cite{rivas2019hybrid}

% Miscelleneous: \cite{bastian2014linkedin}, \cite{gutierrez2019explaining}, \cite{reusens2018evaluating}

% \vspace{1cm}
Recommendation systems find extensive use in a variety of domains such as E-commerce, streaming platforms, due to their ability to enhance user experiences and drive engagement.
They are generated using various techniques, such as collaborative filtering, content-based filtering, or hybrid approaches. In the past, these techniques have been used for the purpose of recommending job opportunities to talent
as well.

Content based recommendations take into account the semantic similarity of attributes that are present in the CV and JDs. There are a number of approaches on how these similarities are calculated. In 
\cite{kessler2012hybrid} Bag of Words method is used, whereas in \cite{bansal2017topic} uses the Latent Dirichlet allocation algorithm. We also find relatively new methods such as word2vec being used for job recommendations \cite{gugnani2020implicit}, \cite{valverde2018job}. %A key takeaway is that the terminology of various attributes is not a one-to-one map \cite{schmitt2016matching}. Same underlying fact can be represented in a multitude of ways which makes recommendations based on such techniques irrelevant in most cases.
%We expect LLM based approach to alleviate this issue.

In Collaborative filtering the recommendations are generated by considering the previous selections a candidate has made and recommending similar jobs in the future.
Memory based collaborative filtering is predominantly used for job recommendations where candidates are grouped based on the similar interactions. Jobs are recommended to candidates by the preference of other candidates belonging to the same group. \cite{lacic2020using} and \cite{lee2017exploiting} apply this method.% for job recommendations.

This is by far the most common approach for job recommendation.
%As mentioned in \cite{de2021job}, ybrid recommeders covering a vast area needs to broken down further into sub units.
Contributions of Model-based methods on shallow embeddings can be found in \cite{chen2017hybrid}.
In recent times Deep neural network based methods have gained a lot of traction with \cite{jiang2020learning}%, \cite{nigam2019job} 
using them for the purpose of job recommendation.
There are also ensemble hybrid methods \cite{de2016scalable} and weighted hybrid which assigns some weightage to the different contributing methods \cite{leksin2017combination}.
Apart from these some notable works in this field include \cite{borisyuk2017lijar}%, \cite{laumer2018job}
which talks about the downside of too many recommendations and the impact of wrong recommendations on a user.

Finally, knowledge based recommendation are based on using inherent knowledge about the job in form of an ontology and it can recommend the jobs to candidates once they too are categorized following the same ontology.
This is discussed in more details in \cite{freire2021recruitment}.%, \cite{gutierrez2019explaining}. 

The LLM based approach circumvents issues related to information extraction and modelling. The Hybrid approach generates predictable recommendations that ensure quality expected by a user, further enriching them by the use of LLMs by capturing attributes that were not captured in the model, by removing bias, and providing justifications.

%%%%%%%%%%%%%%%%%%%%%%%%%%%%%
\section{Conclusion}
%%%%%%%%%%%%%%%%%%%%%%%%%%%%%

Unstructured data is prevalent in job recommendations, as CVs and JDs (JDs) are often in unstructured formats. Converting unstructured data into structured formats has been the traditional approach, but achieving accurate conversion remains challenging. Large language models (LLMs) have emerged as valuable tools in this field. However, LLMs can be slow, expensive, and lack full controllability, resulting in indeterminism and occasional inaccuracies in their results. 
To improve job recommendations, we propose a hybrid approach that combines quantitative attribute filtering followed with LLM analysis. By initially filtering out irrelevant jobs based on quantitative attributes, we can reduce the volume of data before utilizing LLMs to consider the qualitative aspects of the remaining screened jobs. Our experiments demonstrated that relying solely on LLMs may result in job recommendations that lack the required attributes for candidates.
Deterministic methods can also overlook attributes that have not been explicitly modelled and miss important qualitative aspects. In contrast, the hybrid method performs well in terms of efficiency, cost-effectiveness, and capturing both quantitative and qualitative aspects. Additionally, JDs often exhibit bias and highlight only positive qualities. Providing an unbiased perspective of the organization and job position based on the common knowledge of LLMs can empower candidates to make informed decisions.
In conclusion, the LLM-based approach adds value to talent acquisition. It is recommended to be used in conjunction with traditional algorithms to handle scalability and address specific requirements.

%%%%%%%%%%%%%%%%%%%%%%%%%%%%%%%%%%%%%%
\section{Discussion}
%%%%%%%%%%%%%%%%%%%%%%%%%%%%%%%%%%%%%%
%Recommending jobs requires a combination of algorithm advancements, ethical considerations, and ongoing evaluation of recommendations. 

The problem of recommending jobs faces multitude of challenges.
Accurate matching of the skills and requirements of the talent with the job opportunities available is hard. Many JD are vague and poorly defined, making it difficult to identify the right match.
Limited information available about the talent, such as their skills, experience, preferences and complex career goals makes it challenging to recommend personalized job opportunities that align with their qualifications and aspirations.
Keeping up with the evolving job market and ensuring that the recommended opportunities remain relevant is a significant challenge.
Obtaining feedback from talent about the effectiveness of recommendations is essential for continuous improvement. However, gathering feedback requires establishing reliable mechanisms to measure the success and quality of recommendations which is inherently ambiguous.
Further, a real world job marketplace, requires to address complex aspects such as when to recommend, how many recommendations are to be made, how the recommendations influence the user choice, implications of a wrong recommendation and many more.

By fine-tuning prompts, incorporating few-shot learning, training the model on domain-specific data, and leveraging user feedback and reinforcement learning, it is possible to generate more relevant and personalized recommendations in the future.
An important direction of work is to define quality of recommendations unambiguously, possibly using benchmarks. 

%We demonstrate that LLMs are a good fit in some scenarios while not so much in certain other ones. Going forward, the capabilites of LLMs will be augmented by traditional approaches. Also, the  capabilities of LLMs would improve replacing the tasks currently performed by traditional algorithms.

\bibliographystyle{plain}
\bibliography{reco_with_chatgpt_v16}

\end{document}